\documentclass[preprint,3p, authoryear]{elsarticle} %

\usepackage{mathtools}
\usepackage{amssymb}
\usepackage{amsmath}
\usepackage{makecell}
\usepackage{subfig}
\usepackage{xcolor,colortbl}
\usepackage{booktabs}
\usepackage{multirow, makecell, graphicx, array}
\usepackage{hyperref}
\usepackage[capitalize,nameinlink]{cleveref}
\usepackage{xspace}
\usepackage[export]{adjustbox}

\DeclareGraphicsExtensions{.pdf,.png,.eps,.jpg,.tif,.tiff,.ps}

\definecolor{navy}{rgb}{0.1, 0.1, 0.8}
\definecolor{gray}{rgb}{0.4, 0.4, 0.4}
\definecolor{olive}{rgb}{0.1, 0.5, 0.1}
\definecolor{ruby}{rgb}{0.8, 0.1, 0.3}
\definecolor{darkpastelgreen}{rgb}{0.01, 0.75, 0.24}
\definecolor{celestialblue}{rgb}{0.29, 0.59, 0.82}
\definecolor{coral}{rgb}{1.0, 0.5, 0.31}
\definecolor{blue}{rgb}{0.23, 0.44, 0.62}
\definecolor{Goldenrod}{rgb}{0.8,0.8,0}
\usepackage[colorinlistoftodos,prependcaption,textsize=tiny,textwidth=15mm]{todonotes}

\usepackage{soul}
\newcommand{\eat}[1]{{}}

\newcommand{\davidson}{\textsc{Davidson}\xspace} 
\newcommand{\waseem}{\textsc{Waseem}\xspace} 
\newcommand{\reddit}{\textsc{Reddit}\xspace} 
\newcommand{\gab}{\textsc{Gab}\xspace}
\newcommand{\fox}{\textsc{Fox}\xspace}
\newcommand{\mandl}{\textsc{Mandl}\xspace}
\newcommand{\stormfront}{\textsc{Storm\-front}\xspace}
\newcommand{\hateval}{\textsc{Hat\-Eval}\xspace}
\newcommand{\pubfig}{\textsc{Pub\-Figs-L}\xspace}
\newcommand{\pubfigfull}{\textsc{Pub\-Figs}\xspace}

\begin{document}

\newcommand{\titlename}{Generalizing Hate Speech Detection Using Multi-Task Learning: A Case Study of Political Public Figures}
\title{\titlename}

\author[1]{Lanqin Yuan \corref{cor1}}
\ead{lanqin.yuan@student.uts.edu.au}

\author[1]{Marian-Andrei Rizoiu}
\ead{marian-andrei.rizoiu@uts.edu.au}

\cortext[cor1]{Corresponding author}

\affiliation[1]{organization={University of Technology Sydney},
	addressline={15 Broadway, Ultimo NSW 2007},
	postcode={2007},
	city={Sydney},
	country={Australia}}

\begin{abstract}
Automatic identification of hateful and abusive content is vital in combating the spread of harmful online content and its damaging effects. Most existing works evaluate models by examining the generalization error on train-test splits on hate speech datasets. These datasets often differ in their definitions and labeling criteria, leading to poor generalization performance when predicting across new domains and datasets. This work proposes a new Multi-task Learning (MTL) pipeline that trains simultaneously across multiple hate speech datasets to construct a more encompassing classification model. Using a dataset-level leave-one-out evaluation (designating a dataset for testing and jointly training on all others), we trial the MTL detection on new, previously unseen datasets. Our results consistently outperform a large sample of existing work. We show strong results when examining the generalization error in train-test splits and substantial improvements when predicting on previously unseen datasets. Furthermore, we assemble a novel dataset, dubbed \pubfigfull, focusing on the problematic speech of American Public Political Figures. We crowdsource-label using Amazon MTurk more than $20,000$ tweets and machine-label problematic speech in all the $305,235$ tweets in \pubfigfull. We find that the abusive and hate tweeting mainly originates from right-leaning figures and relates to six topics, including Islam, women, ethnicity, and immigrants. We show that MTL builds embeddings that can simultaneously separate abusive from hate speech, and identify its topics.
\end{abstract}

\maketitle

\section{Introduction}
\label{sec:introduction}
With the increasing prevalence of online media platforms in our day-to-day lives, detecting hateful and abusive content has become necessary to prevent the pollution of online platforms by problematic and malicious users~\citep{Schneider2023}. 
Automatic detection of such harmful content has recently received significant attention from the research community. 
Currently, most existing works evaluate their models \emph{in context} using train-test splits: the model is sequentially trained and then tested on the same dataset.
However, several recent works \citep{10.1145/3331184.3331262, swamy-etal-2019-studying,yin2021generalisable,FORTUNA2021102524} raised concerns over the poor generalization performance of such existing models when applied to hate speech datasets other than those used to train the model. 
This poor performance persists even for datasets gathered from the same platform. 

\begin{figure}[tbp]
	\centering	
	\includegraphics[width=.8\columnwidth]{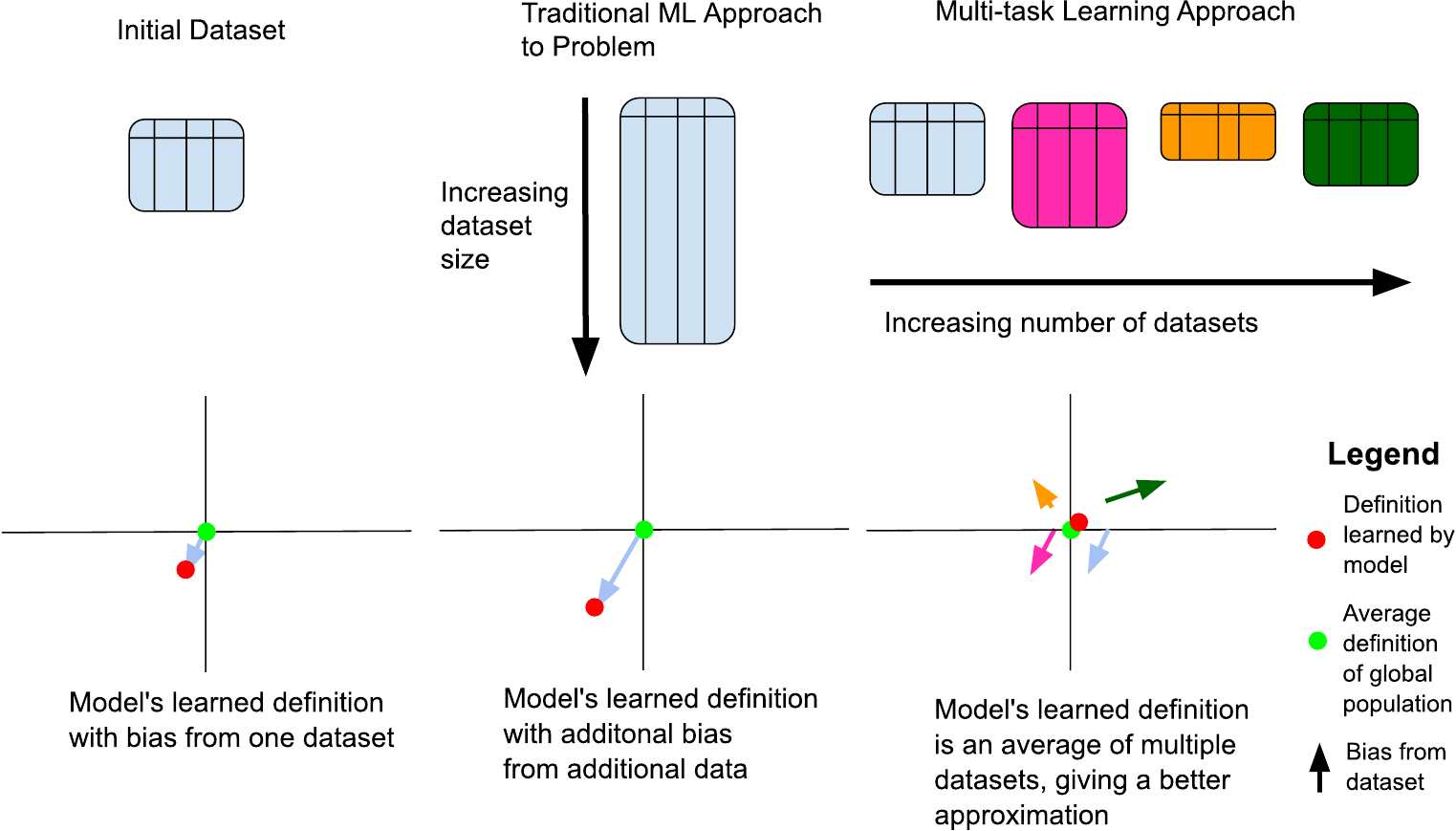}	
	\caption{
		Different approaches to addressing labeling bias in hate speech datasets.
		The traditional Machine learning approach increases the size of the training dataset by adding more labeled rows with the same labeling definition, leading to additional bias to that labeling criteria. 
		Our novel multi-task learning approach allows for increasing the number of datasets and definitions in the training pipeline for a more general representation.
	}
	\label{fig:increasing_dataset}
\end{figure} 
A key challenge in building generalizable models is the lack of a universally agreed-upon definition of hate speech that is specific enough to be operationalized.
There are many hateful and abusive speech facets (dubbed in this work as \emph{domains}), such as racism, sexism, ableism, bullying, harassment, incitement of violence, and extremism.
Most prior works concerned with hate speech detection concentrate on specific domains, which translates to differences in the labeling criteria.
As a result, each dataset captures only a fraction of the hate speech, partially explaining why models trained on single datasets generalize poorly to other datasets concerned with other domains.

\textbf{Problem Statement.}
The work presents a method for training hate speech detection models that account for multiple definitions of hate speech.
Given the absence of a universally agreed-upon definition, existing works leverage machine learning models to estimate what constitutes hate speech.
Such models learn from labeled data what represents hate speech, therefore shifting the difficulty from defining hate speech to assembling representative datasets.
Due to how they are built---typically using human labelers in a narrow domain--such datasets will always suffer from sample size and labeling biases.
As a result, the trained hate speech classifiers are unlikely to generalize to new domains and datasets.

\textbf{Research Questions.}
We address the problem by examining two open questions concerning hate speech detection. 
The first research question relates to constructing models that account for the various definitions of hate across different datasets.
Most existing work adopt a narrow definition, construct labeled datasets, and measure the generalization error on train-test splits \citep{DBLP:journals/corr/DavidsonWMW17, waseem-hovy-2016-hateful, basile-etal-2019-semeval, de-gibert-etal-2018-hate}.
Given the discussion in the previous paragraph, the results are overly optimistic and fail to generalize to new domains \citep{10.1145/3331184.3331262, swamy-etal-2019-studying}. 
The question is \textbf{can we construct a model that can utilize multiple hate speech datasets to capture the differing definitions of hate across different datasets to improve a model's classification performance?}
The second question concerns detecting hate speech in previously unseen domains and datasets.
A limited number of works attempted to train on multiple models \citep{Yuan2023,Waseem2018} but generally do not study predictivity on previously unseen datasets and domains.
The research question is \textbf{can we build a detection approach that generalizes to an entirely new dataset containing hateful and abusive speech?}

\textbf{Our Solution.}
Addressing the first question, we contribute a novel transfer learning pipeline that accounts for the multiple facets of hate and abusive speech by training on multiple datasets in parallel.
Intuitively, this reduces the bias and increases the generalization of the constructed detection model.
\cref{fig:increasing_dataset} ilustrates this process.
Any labeled dataset (top-left panel of \cref{fig:increasing_dataset}) embeds a degree of bias in defining hate speech compared to the global population -- all possible types of hate speech at a given moment.
This bias is schematically shown dimensionality-reduced in the bottom-left panel.
The typical but naive solution to increasing the performance of hate speech detection is to increase the size of the training dataset (top-middle panel).
However, as the same criteria for constructing the dataset are applied, this results in a similar or even more embedded bias (bottom-middle panel), as the model overfits to a single definition of hate speech.
Our proposed method learns from multiple labeled datasets (top-right panel), each with its own hate speech domain and labeling criteria.
These datasets have biases (arrows in the bottom-right panel), which get canceled and averaged out as our model learns a single, more encompassing definition of hate speech.
We adopt a \textit{Multi-Task Learning} (MTL) technique with Hard Parameter Sharing \citep{baxter_bayesianinformation_1997} to the task of hate speech detection. 
We train on eight publicly available datasets (as opposed to prior literature, which uses 2 or 3 datasets). 
Furthermore, we construct a novel dataset containing social media postings of American public political figures human-annotated for hate and abuse speech\footnote{We publicly release the trained model, the detection and training code, and the American Political Figures dataset, which are available at \url{https://code.research.uts.edu.au/14386080/mtl\_hatespeech}}. %
We fine-tune a single BERT language model \cite{DBLP:journals/corr/abs-1810-04805} to which we attach as many classification heads as datasets to train.
Each classification head is adapted to detect the hate classes specific to its datasets; however, the gradients are back-propagated into a single language model.
Intuitively, this constructs a single representation that captures a more generic definition of hate.
Our approach differs from existing works in our pipeline architecture and our utilization of MTL across a significantly higher number of hate speech datasets.
Our model produces strong predictive performances in train/test split scenarios;
we compare MTL against nine state-of-the-art hate speech detection architectures reported on several datasets.
MTL outperforms the competing approaches (12 wins, 1 draw, 7 losses) on their reported performance measures.

We address the second open question in a sequence of three steps.
In the first step, we employ a leave-one-out scheme to evaluate unseen hate speech detection.
We train MTL on all but the target dataset and evaluate on the target dataset that MTL never saw during training.
MTL obtains the best prediction performances compared to existing literature, except for datasets whose labeling is so specific that using a generalized model hurts performances. 
In the second step, we use MTL to study the generalization of classifiers trained on individual datasets to new unseen datasets.
We find that specific pairs of datasets have high mutual generalization -- typically those proposed by the same authors in the same work.
This confirms that the dataset construction and labeling reflect the author-specific definition of hate speech.
We also note that MTL generalizes best for 7 (out of 9) datasets. 

The third and final step is to test hate speech detection in a novel, previously unexplored hate speech domain using MTL.
We construct a brand new labeled hate speech Twitter dataset of $305,235$ tweets from 15 American public figures across both left and right political leanings, dubbed the \pubfigfull dataset. 
We use Amazon Mechanical Turk crowdsourcing to label a subset of $20,327$ tweets -- dubbed the \pubfig dataset. 
To our knowledge, this is the first dataset focusing on the hate speech of American public figures. 
Our dataset has previously unexplored particularities, such as dealing with covert hate speech and focusing on a small number of public political figures instead of many anonymous users on an online social media platform. 
We apply the proposed MTL pipeline model to the \pubfig dataset to examine the posting behavior of the figures in the dataset. 
We uncover that right-leaning figures in our dataset post more inappropriate content than left-leaning figures and identify that hateful and abusive speech primarily concentrates on 6 topics: Islam, Women, Race and Ethnicity, Immigration and Refugees, Terrorism and Extremism, and American Politics.
We examine the effects of MTL training on the BERT embedding space, finding that MTL training increases the distinctness of hate speech and abusive content in the embedding space, both from neutral content and from each other.
We further examine the distinctness of particular facets of hate, finding that misogyny and Islamophobia, in particular, became significantly more distinct in the space compared to other facets of hate.

\textbf{The main contributions of this work are as follows:}
\begin{itemize}
    \item A Multi-Task Learning pipeline using multiple datasets to construct a more encompassing representation of hate speech;
    \item An extensive analysis of the generalization of hate speech detection to new, unseen datasets and domains;
    \item A novel Twitter dataset containing posts from 15 American public political figures, annotated for covert hate and problematic speech.
\end{itemize}

\section{Background}
\label{sec:background}

This section discusses
the definitions of hate speech, related works in hate speech classification, and transfer learning approaches.

\textbf{Defining Hate Speech.}
Hate speech is not easily quantifiable as a concept \citep{MacAvaney2019,Kong2022,Kong2020a,Kong2021}. 
It lies on a continuum with offensive speech and other abusive content such as bullying and harassment.
Some definitions given in the literature are as follows: 
The United Nations~\citep{noauthor_united_2019} defines hate speech as ``any kind of communication in speech, writing or behavior, that attacks or uses pejorative or discriminatory language concerning a person or a group based on who they are, in other words, based on their religion, ethnicity, nationality, race, color, descent, gender or other identity factor''.
\citet{DBLP:journals/corr/DavidsonWMW17} defines hate speech as ``language that is used to express hatred towards a targeted group or is intended to be derogatory, to humiliate, or to insult the members of the group". 
\citet{Fortuna10.1145/3232676} surveyed definitions of hate speech and produced their own: ``Hate speech is language that attacks or diminishes, that incites violence or hate against groups, based on specific characteristics such as physical appearance, religion, descent, national or ethnic origin, sexual orientation, gender identity or other, and it can occur with different linguistic styles, even in subtle forms or when humor is used". 
In contrast, \citet{schmidt-wiegand-2017-survey} defines a much broader definition of hate speech as an umbrella term for all content on the hateful to abusive continuum.
Most of these definitions are enumerations of the facets of hate speech, which make it difficult to transform into detection.
These definitions, while similar, have subtle differences in their semantics which can cause ambiguity in dataset construction and detection.

\textbf{Hateful and Abusive Speech Classification.}
Early approaches for hate speech classification utilized non-neural network-based classifiers, usually in conjunction with manual feature engineering. 
Examples of such works are \citet{DBLP:journals/corr/DavidsonWMW17} and \citet{waseem-hovy-2016-hateful} which used  
various engineered features in conjunction with a logistic regression classifier. 
More recently, \citet{MacAvaney2019} utilized a Multi-view SVM with feature engineering,
reporting results similar to modern neural network models.

Recent advances in deep learning have seen the state-of-the-art dominated by deep neural network-based models. 
Initial models utilized recurrent or convolutional neural networks \citep{10.1007/978-3-319-93417-4_48} with textual features, often in conjunction with non-neural classifiers \citep{Badjatiya2017} and feature engineering \citep{Agrawal2018,Rizoiu2016}.

The introduction of large pre-trained transformer language models such as the ``Bidirectional Encoder Representations from Transformers'' (BERT) \cite{DBLP:journals/corr/abs-1810-04805}, and its variants have shown impressive performance in several NLP-related tasks, including hate speech detection \citep{mozafari_bert-based_2019,Madukwe_GA_BERT,swamy-etal-2019-studying, ROY2022101386}. 
Our work extends upon these works by exploring transfer learning and multi-task learning in conjunction with transformer-based models.

\textbf{Transfer and Multi-Task Learning.}
Transfer learning is the exploitation of knowledge gained in one setting to improve the generalization performance in another setting \citep{10.5555/3086952}. 
Formally, given source domain $D_S$ and source task $T_S$, target domain $D_T$ and target task $T_T$ where $D_S \neq D_T$ or $T_S \neq T_T$, transfer learning seeks to make an improvement to the learning of the target predictive function $f_T(\cdot)$ in $D_T$ using knowledge in $D_S$ and $T_S$ \citep{5288526}.

The most common form of transfer learning is Sequential Transfer Learning (STL): the model is trained on related tasks one at a time, then fine-tuned to adapt the source knowledge to the target domain. 
Multi-Task Learning (MTL) is an alternative paradigm known as parallel transfer learning. 
MTL seeks to transfer knowledge between several target tasks simultaneously and jointly, rather than sequentially towards a single target task \citep{baxter_bayesianinformation_1997}. 
The tasks act as regularizers for each other in the joint model. 

\textbf{Closest Related Works.}
Several existing works relate closely to our work. 
We discuss these works in two groups.

The first group examines the performance of hate speech detection models in cross-dataset settings, and hate speech classification in previously unseen hate speech datasets. 
\citet{Guimaraes} investigated the composition, vocabulary and targets of hate in several hate speech datasets, drawing attention to the conflicting definitions of hate speech contained within the datasets. 
They examined the cross-dataset classification performance between 6 different hate speech datasets.
Similar to our work, they fine-tuned a BERT classification model on a single dataset and then tested on the other five datasets.
They find that datasets with closer definitions of hate speech and similar compositions in the type of hate speech tend to achieve better cross-dataset performance with each other.
Our work differs in our employment of MTL to transfer knowledge between datasets and as a means to improve the cross-dataset generalization performance.
\citet{Chiril2021EmotionallyIH} explored the ability of hate speech detection models to transfer knowledge from generic hate speech datasets to more granular topic-specific hate speech detection tasks. 
They explore two evaluation schemes: the first scheme trains on a single topic-general hate speech dataset and then tests on one of several topic-specific datasets.
The second scheme concatenates the training set of all topic-specific datasets and uses it to train a single model. 
They examined various classification models based on LSVM, LSTM, CNN, ELMo, and BERT. 
Our work extends upon transferring knowledge from multiple datasets by exploring multi-task learning over concatenation to learn a generalized representation implicitly.

The second group of existing literature examines the application of Multi-Task Learning for detecting hateful and abusive speech.
Various studies have explored MTL's utility in this area. 
\citet{PlazaDelArco} harnessed MTL for hate speech detection, integrating multiple detection tasks within polarity and emotion knowledge classification to augment the hate speech classifier. 
Our study sets itself apart by applying MTL across disparate datasets instead of varying tasks within the same dataset. 
\citet{Waseem2018} employed MTL across multiple datasets, utilizing hard parameter sharing within a Recurrent Neural Network classification framework. 
\citet{Yuan2023} adopted an MTL strategy, generating generalized embeddings via a bi-directional LSTM model across two datasets. \citet{KAPIL2020106458} investigated multiple MTL configurations using five datasets with distinct labels related to hate speech. Unlike these studies, our approach leverages a BERT architecture, diverging from more traditional neural network models like LSTMs and RNNs. 
We incorporate a broader array of datasets, and, most importantly, we concentrate on classifications in previously unseen datasets.

\citet{Ghosh_Priyankar_Ekbal_Bhattacharyya_2023} proposed an MTL framework that tackles hate speech detection along with aggressive posting identification, emotion detection and several other related sub-tasks in offensive speech detection.
Our methodology differs significantly in the MTL architecture utilized. 
We employ a shared BERT unit across various datasets, accompanied by a single classification head for each dataset. 
In contrast, \citet{Ghosh_Priyankar_Ekbal_Bhattacharyya_2023} utilized two independent neural network channels, each dedicated to a dataset, along with a shared XLMR encoder. This encoder merges with the channels before connecting to multiple classification heads, one for each task.
\citet{PlazaDelArco} explored a MTL configuration to enhance hate speech classification on Twitter in Spanish, utilizing auxiliary sentiment and emotion classification tasks to aid the hate speech detection task. 
Unlike their approach, our research does not leverage auxiliary tasks. 
Instead, we train across multiple hate speech datasets with varying definitions of hate, aiming to derive a generalized hate speech representation to bolster predictions on new, unseen datasets.

\section{Methodology}
\label{sec:methodology}
This section details our MTL pipeline (\cref{subsec:model}), its training (\cref{subsec:training_pipeline}), and the classification of unseen datasets (\cref{subsec:classification_on_unseen_data}).

\begin{figure}[tbp]
	\centering	
	\includegraphics[width=.95
	\columnwidth]{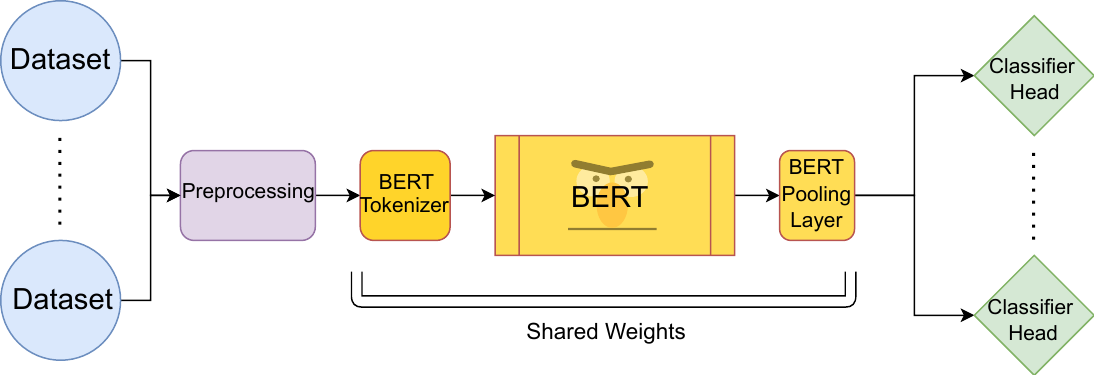}	
	\caption{
		Schema of the Multi-Task Learning pipeline. 
		An arbitrary number of datasets	are used to train a single model jointly. 
		Each dataset-specific classification head propagates its loss through a single shared BERT unit to produce a generalized representation of hate speech.
	}
	\label{fig:bert_schema}
\end{figure} 
\subsection{Model}
\label{subsec:model}

\cref{fig:bert_schema} shows the schema of our proposed BERT-based \citep{DBLP:journals/corr/abs-1810-04805} multi-task transfer learning pipeline. 
It consists of the following: 

\textbf{The Preprocessing Unit} standardizes the text input by removing capitalization, repetitive punctuation, redundant white spaces, emojis, and URLs in text. 
It uses a single space character to separate all words and punctuation.
We filter out empty sequences or only containing the Twitter-specific retweet flag ``RT'' after preprocessing.

\textbf{The BERT Unit} consists of the pre-trained BERT tokenizer, the pre-trained BERT language model, and a pooling layer. 
The BERT tokenizer creates a sequence of tokens as input for the BERT model from the preprocessed text. 
The BERT model creates a representation for each token in the sequence. 
The BERT pooling layer constructs a fixed-sized sentence embedding by pooling together the individual token representations.
It takes inputs from the BERT token outputs and the hidden state of the last BERT layer after processing the first token of the sequence (the ``[CLS]" token). 
The pooling layer consists of a fully connected layer and a $tanh$ activation function, formally defined as $$\tanh(z) = \frac{e^{z} - e^{-z}}{e^{z} + e^{-z}}$$ where $z = \vec{w}^T \vec{x}$ given inputs $\vec{x}$ and weights $\vec{w}$.
We use the original pre-trained weights (\textit{bert-base-uncased}) \citep{DBLP:journals/corr/abs-1810-04805} to initialize the tokenizer, the BERT model and the pooling layer. 
During pretraining, the Linear layer weights are trained using the next sentence prediction (classification) objective. 

\textbf{The Dataset-Specific Classifier Heads} take the representation produced by the pooling layer and produce a classification for each task (corresponding to each dataset).
Each head consists of a 3-layer neural network with a hidden size of 512 and softmax activation, formally given as $$\text{Softmax}(\vec{z})_i = \frac{e^{z_i}}{\sum_{j=1}^{K} e^{z_j}}$$ for the $i$-th class of $K$ classes where $\vec{z} \in \mathbb{R}^K$ is the logit output from the hidden layer. 
The classification heads do not share weights, producing different predictions based on the single set of BERT encodings.
This incentivizes BERT to construct representations useful to all classifier heads.

\subsection{Training Pipeline Implementation Details}
\label{subsec:training_pipeline}

\textbf{Dataset Preprocessing.}
The training constructs a task-agnostic representation of hateful and abusive speech by jointly fine-tuning BERT's parameters using several datasets.
The datasets can have arbitrary numbers of classes (see \cref{tab:datasets}).
Hate speech datasets are known to have a heavy class imbalance~\citep{Yuan2023,madukwe-etal-2020-data}; 
therefore, we use a stratified random split of 8:1:1 (training : validation : test set) to ensure that all subsets contain the same ratio of classes.
We further use random oversampling of the minority classes during training.

\textbf{Training and Loss Function.}
We train the MTL pipeline for $10$ epochs with a learning rate of $2e-5$. 
Each epoch consists of a pass over each of the $n$ datasets' training sets.
We use mini-batching with a batch size of $512$. 
Once all heads have completed the epoch, losses are accumulated and summed over all classification heads, then propagated using the Adam optimizer \citep{kingma2017adam}.
This optimize for their own respective dataset classification task while the shared BERT must optimize for all datasets classification tasks simultaneously. 
results in the individual classification heads having different loss to the shared BERT unit.

The loss for each dataset classification head is defined as its cross-entropy loss, formally given for dataset $d_i$ with labels $y_i$ and predictions $\hat{y}_i$ as:
\begin{equation*}
H(y_i, \hat{y}_i) = -\sum_{j\in d_i} y_{ij} \cdot \log(\hat{y}_{ij})
\end{equation*}

The total BERT loss at each epoch is the sum of the individual loss for all $K$ datasets. Formally this is therefore defined as:

\begin{align*}
	Loss &= - \sum_{i=1}^{K} H(y_i, \hat{y}_i)  \\
	&= - \sum_{i=1}^{K} \sum_{j\in d_i} y_{ij} \cdot \log(\hat{y}_{ij})
\end{align*}

\textbf{Computational Complexity.}
The MTL approach is very computationally efficient.
Unlike Sequential Transfer Learning, which requires learning from the different tasks sequentially, one after the other (see \cref{sec:background}), the Parallel Transfer Learning paradigm that we employ in MTL allows the concurrent computation of each dataset classification head.
Consequently, all dataset computations can be achieved in parallel with sufficient computational resources (i.e., GPUs).
As a result, the computation time increase for adding new datasets is only marginal and relates mainly to initialization and data transfer.
The update of the shared model requires completing all classification heads' processing before its backward pass, but additional datasets do not increase the overall computational footprint.
This results in a training runtime complexity that scales linearly with the number of datasets.

\textbf{Selecting Best Model.}
We evaluate using the holdout validation set at each epoch. 
We select the model weights with the highest validation macro-F1 score over all epochs as the final model weights. 
We also explored selecting the final weights using the validation loss but found the performance worse than macro-F1.

\textbf{Single Dataset Baselines} 
use the same architecture as MTL, but with only one input dataset and one classification head.
Therefore, we tune the BERT and classifier on a single task.
We use the Single Dataset Baseline in \cref{subsec:unseen_data_classification} to evaluate dataset pairwise generalization.

\subsection{Classification on Unseen Datasets}
\label{subsec:classification_on_unseen_data}
Each classification head of the MTL pipeline independently predicts an arbitrary number of problematic classes, depending on each dataset's labeling (see \cref{tab:datasets}).
As the number of classes in new, previously unseen datasets is unknown, we construct a binary classification of content as problematic/harmless.
We build a binary mapping for each dataset by joining the classes shown in red italic font in \cref{tab:datasets} into a single problematic class (and the others into the harmless class).

We propose two schemes for classifying unseen datasets: \emph{New Classification Head} (NCH) and \emph{Majority Vote} (MV). 
NCH trains a new head on the binarized versions of the available datasets.
First, we build a new training set that concatenates training instances from all datasets.
Second, we build a validation set that concatenates all datasets' validation and testing sets.
Finally, we freeze the MTL-tuned BERT and train a new binary classifier head for 10 epochs, selecting the final weights based on the best validation performance.
MV leverages the trained dataset-specific classifier heads.
Each classifier makes an individual prediction -- binarized using its specific dataset mapping.
A majority vote of classifiers selects the final classification label for each instance in the unseen dataset.

NCH and MV differ in the effect of dataset sizes on the final classification. 
In the NCH scheme, each dataset contributes proportionally to its size, as more training information originates from larger datasets than smaller ones. 
By contrast, the MV scheme gives smaller datasets equal weight to larger datasets in the majority vote. 
We investigate both schemes in \cref{sec:experiments_and_results}. 
For single dataset baselines (see \cref{subsec:training_pipeline}), we only binarize the output of the classification head -- i.e., the MV scheme with a single voter.

\section{Datasets} 
\label{sec:existing_datasets}

\begin{table}[t]
    \centering
    \caption{
        \textbf{The datasets used in this work with the number of labeled examples.}
        Datasets contributed from this work are highlighted in \textbf{bold}.
        \textcolor{red}{\textit{Problematic}} classes are shown in \textcolor{red}{red} italics.
        Note that the \pubfigfull is not manually labeled, and only the sample \pubfig is labeled for problematic speech.
    }
    \label{tab:datasets}
       \small
    \setlength{\tabcolsep}{2pt}
    \begin{tabular}{lrrrr}
        \toprule
        \textbf{Dataset} & \textbf{Classes}                                                                                        & \textbf{\#Neutral} & \textbf{\#Problematic} & \textbf{\#Total} \\
        \midrule
        {\davidson} \citep{DBLP:journals/corr/DavidsonWMW17}      & Neither, \textbf{\textit{\textcolor{red}{Offensive}}}, \textbf{\textit{\textcolor{red}{Hate}}}          & $4,162$             & $20,620$            & $24,782$          \\%

        {\waseem} \citep{waseem-hovy-2016-hateful}        & Neither, \textbf{\textit{\textcolor{red}{Racism}}}, \textbf{\textit{\textcolor{red}{Sexism}}}           & $11,501$            & $5,406$             & $16,907$          \\%

        {\reddit} \citep{DBLP:journals/corr/abs-1909-04251}       & Non-Hate, \textbf{\textit{\textcolor{red}{Hate}}}                                                       & $10,053$            & $3,130$             & $13,183$          \\%

        {\gab} \citep{DBLP:journals/corr/abs-1909-04251}          & Non-Hate, \textbf{\textit{\textcolor{red}{Hate}}}                                                       & $15,111$            & $11,046$            & $26,157$          \\%

        {\fox} \citep{gao-huang-2017-detecting}          & Non-Hate, \textbf{\textit{\textcolor{red}{Hate}}}                                                       & $919$               & $332$               & $1,251$           \\%

        {\mandl} \citep{10.1145/3368567.3368584}        & Neutral, Profane, \textbf{\textit{\textcolor{red}{Hate}}}, \textbf{\textit{\textcolor{red}{Offensive}}} & $3,135$             & $1,883$             & $5,018$           \\%

        {\stormfront} \citep{de-gibert-etal-2018-hate}    & Non-Hate, \textbf{\textit{\textcolor{red}{Hate}}}, Skip/Unclear, Relation & $9,330$             & $1,089$             & $10,419$          \\%

        {\hateval} \citep{de-gibert-etal-2018-hate}      & Non-Hate, \textbf{\textit{\textcolor{red}{Hate}}}                                                       & $6832$              & $4,926$             & $11,758$          \\%
		\midrule
        \textbf{\pubfig}        & Neutral, \textbf{\textit{\textcolor{red}{Abuse}}}, \textbf{\textit{\textcolor{red}{Hate}}}              & $17,963$            & $2,364$             & $20,327$          \\%
        \textbf{\pubfigfull}        & -- & -- & -- & $305,235$          \\%
        \bottomrule
    \end{tabular}%
\end{table} 
In this section, we discuss the eight publicly available datasets that we use in this work. 
We summarize them in \cref{tab:datasets}, together with their number of instances and classes, highlighting the ones we deem problematic.
The rest of this section details each dataset; the following section (\cref{sec:public_figures_twitter_dataset}) introduces the novel datasets we construct -- dubbed \pubfigfull.

\davidson~\citep{DBLP:journals/corr/DavidsonWMW17} is a widely used Twitter dataset in hate speech classification and other related applications \citep{Badjatiya2017,mozafari_bert-based_2019,10.1007/978-3-319-93417-4_48}. 
The dataset explicitly differentiates between hateful speech and offensive speech.
The dataset was constructed by first collating a list of potentially hateful tweets by conducting a keyword search using words and phrases from Hatebase.org. Crowdsourced workers from CrowdFlower then labeled a sample of $25,000$ tweets.
Workers were provided with definitions of the classes and asked to consider the entire context of the text rather than focusing on individual words.
Workers were told that the presence of particular terms, however offensive they may be, does not indicate hate speech.
At least three workers labeled each data instance.

\waseem~\citep{waseem-hovy-2016-hateful} is a Twitter dataset widely used in prior works.
It focuses exclusively on two specific facets of hate: racism and sexism.
The authors collated a corpus of tweets based on a keyword set that contains frequently occurring terms in hateful tweets and references to specific entities, such as TV shows which often incite racist and sexist tweets.
The authors added the tweets of manually identified prolific posters of hateful content to this corpus.
The labeling was done by the two authors and an outside annotator.
Several existing works, including \citet{Waseem-2016-racist} himself, have raised questions regarding the quality of the annotations in the dataset. 
\citet{10.1145/3331184.3331262} highlight the skewed distribution of users who contribute to the \textit{racist} and \textit{sexist} classes, with 8 users accounting for all tweets labeled as \textit{racist}.
	
\reddit~\citep{DBLP:journals/corr/abs-1909-04251}: 
Reddit is a well-known social media platform comprising subreddits, where user-created communities discuss topics or themes.
The \reddit dataset was collected from ten subreddits based on their tendency to contain toxicity and hate speech, such as \texttt{r/DankMemes}, \texttt{r/MensRights}, and \texttt{r/TheDonald}.
The top 200 posts of each subreddits sorted by Reddit's ``Hottest'' ordering were collated. 
The conversation threads were filtered using the hate keywords from \citet{DBLP:journals/corr/abs-1804-04649} and annotated using Amazon Mechanical Turk as either hateful or non-hateful; three different workers annotated each thread.
	
\gab~\citep{DBLP:journals/corr/abs-1909-04251}:
Gab is a social media and microblogging platform with functionalities similar to Twitter, well known for its far-right user base \citep{doi:10.1177/14614448211024546}. 
The \gab dataset originates from the same work as the Reddit dataset and uses a similar methodology.
The authors used hateful keywords \citep{DBLP:journals/corr/abs-1804-04649} to identify potentially hateful posts, which are then labeled as hateful or non-hateful using Amazon Mechanical Turk.
Each post was assigned to 3 different workers.

\fox~\citep{gao-huang-2017-detecting} is a small binary -- hate speech or not -- dataset that contains user comments from ten news discussion threads on the Fox News website.
Four of the ten threads were annotated by two native English speakers, who discussed the labeling criteria before annotation. 
The other six threads were annotated by only one of the two annotators.
	
\mandl~\citep{10.1145/3368567.3368584} is a Twitter dataset released as a part of the HASOC track of the 2019 ACM Forum for Information Retrieval Evaluation (FIRE) conference. 
We only use the English dataset. 
Labeling was done in two steps. 
First, tweets were labeled as \textit{hateful and offensive} or \textit{not hateful or offensive}. 
\textit{Hateful and offensive} tweets were then re-examined and further divided into 3 classes, resulting in a final set of 4 classes: \textit{Neutral}, \textit{Hate}, \textit{Offensive}, and \textit{Profane}. 
For this work, we consider only the classes \textit{Hate} and \textit{Offensive} as problematic;
according to most definitions of hate speech (see \cref{sec:background}), profanities (while displeasing) cannot be considered problematic content.
``Several juniors'' performed the annotations after being given rough guidelines and definitions.
	
\stormfront~\citep{de-gibert-etal-2018-hate}: 
Stormfront is a neo-Nazi Internet forum considered one of the major racial hate hubs.
The \stormfront dataset contains $9,916$ sentences from $500$ posts posted across $22$ sub-forums on the Stormfront website between 2002 and 2017. 
Three annotators first labeled $1,144$ sentences as either \textit{hate}, \textit{no hate}, or \textit{skip/unclear}.
A \textit{relation} label is also used for posts hateful given the conversation chain's full context, but not by themselves.
Our work treats each sentence as an individual data instance. 
Hence, we consider the \textit{skip/unclear} and \textit{relation} classes non-hate. 
	
\hateval~\citep{basile-etal-2019-semeval} is a Twitter dataset containing tweets from July to September 2018. 
The \textit{hate speech} class in this work is focused explicitly on hate against immigrants and women. 
The original work contributes a Spanish dataset in addition to the English one, and it provides additional labels for the target of hate and the aggressiveness of the content. 
We do not use these in this work.

\begin{table*}[]
	\caption{
		\textbf{The American Political Figures captured in the \pubfig dataset.}
		Counts of Neutral, Abusive and Hate tweets from each figure, alongside their brief description.
	}
	\small
	\setlength{\tabcolsep}{2pt}
	\begin{tabular}{lllllll}%
		\toprule
		\textbf{Figure} & \renewcommand\cellalign{ll}\makecell{\textbf{Perceived} \\ \textbf{Political} \\ \textbf{Leaning}} & \#Neutral & \#Abuse & \#Hate & \#Total & \textbf{Description} \\
		\midrule
		AJ & Right & 1666 & 182 & 173 & 2021    & Far right show host and prominent conspiracy theorist. \\
		MG & Right & 486 & 45 & 27 & 558        & American Republican politician. Georgia Congress repr. \\
		CO & Right & 261 & 38 & 21 & 320        & American conservative influencer and political commentator. \\
		AK & Right & 1086 & 61 & 23 & 1170      & Right winged American political activist. \\
		AC & Right & 5143 & 523 & 464 & 6130    & American conservative media pundit and author. \\
		LI & Right & 1474 & 28 & 10 & 1512      & American conservative television host. \\
		BS & Right & 1753 & 129 & 56 & 1938     & American conservative political commentator. \\
		DT & Right & 1972 & 197 & 85 & 2254     & Former Republican US president.\\
		DJ & Right & 1862 & 166 & 56 & 2084     & Relative of DT.\\
		TS & Left  & 247 & 1 & 0 & 248          & Popular American Musician.\\
		BR & Left  & 282 & 2 & 0 & 284          & American Democrat politician. \\
		BO & Left  & 76 & 0 & 0 & 76            & Former Democrat US president. \\
		MO & Left  & 34 & 0 & 0 & 34            & Former US first lady. \\
		AO & Left  & 354 & 4 & 4 & 362          & American Democrat politician. New York Congress repr.\\	
		IO & Left  & 1267 & 46 & 23 & 1336      & American Democrat politician. Minnesota Congress repr.\\
		\bottomrule
	\end{tabular}
	\label{tab:figures_table}
\end{table*}

\section{The \pubfigfull Twitter Dataset}
\label{sec:public_figures_twitter_dataset}
\begin{figure}[t]
	\centering
	\newcommand\myheight{0.50}
	\includegraphics[height=\myheight\textheight]{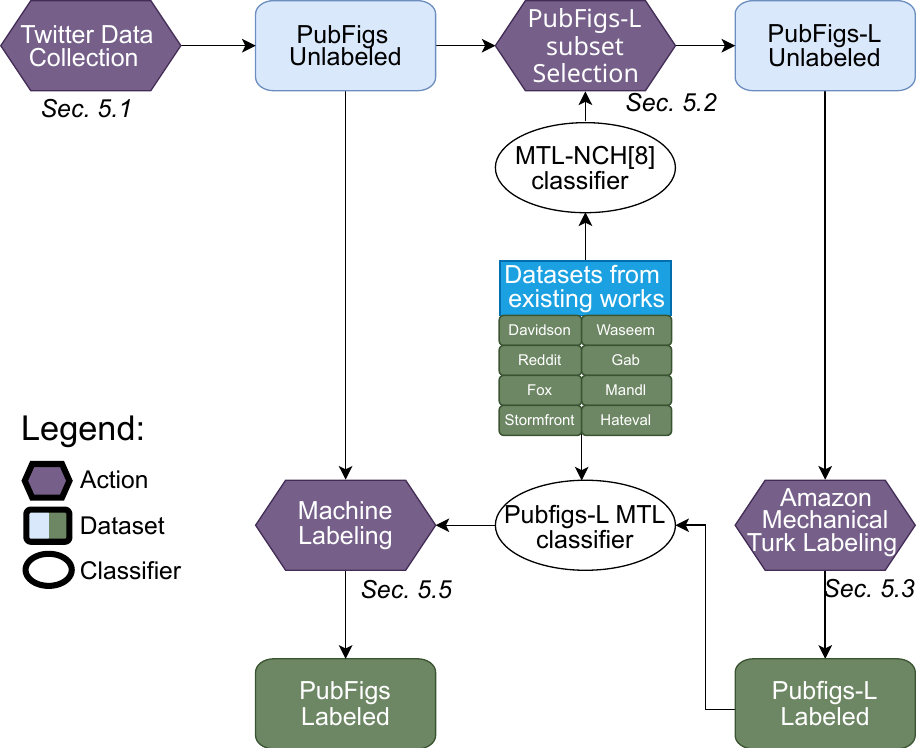}%

	\caption{
		The construction process of \pubfigfull and \pubfig datasets. 
		We detail each action (\textcolor{purple}{purple hexagon}) in \cref{sec:public_figures_twitter_dataset}:
		Twitter Data Collection in \cref{subsec:pubfig-collection},
		\pubfig subset Selection in \cref{subsec:subset-selection},
		Amazon Mechanical Turk Labeling in \cref{subsec:mturk_labelling}, and
		Machine Labeling in \cref{subsec:full-pubfig}.
	}
	\label{fig:pubfigs_construction_diagram}%
\end{figure}

This section presents the construction of the \pubfigfull dataset we contribute and its human-labeled subset \pubfig. 
\cref{fig:pubfigs_construction_diagram} schematically shows the construction process of \pubfigfull and \pubfig.
Tweets are first collected from Twitter to construct the unlabeled \pubfigfull (details in \cref{subsec:pubfig-collection}). 
Next, we use the MTL classifier to construct a subset that we manually label using Amazon Mechanical Turk, giving us \pubfig (\cref{subsec:subset-selection}).
Finally, we train a new classifier with \pubfig added into the MTL training process, which we then use to machine-label \pubfigfull (\cref{subsec:full-pubfig}).

\subsection{Twitter Data Collection}
\label{subsec:pubfig-collection}
We gather historical tweets from 15 American public figures -- such as former presidents, conservative politicians, far-right conspiracy theorists, media pundits, and left-leaning representatives perceived as very progressive.
The dataset mainly pertains to public figures directly associated with American politics with a social media following. 
Some figures are not directly involved in American politics but have a significant public or social media following.
We select figures based on their perceived conservative (right-) vs. liberal (left-) political leaning and media presence to cover a range of personalities and social media behavioral patterns. 
We collected $305,235$ tweets from 16 Twitter accounts (one figure used two accounts with similar amounts of activity). 
\cref{tab:figures_table} outlines the figures in the dataset, with each figure only being referred to by a pseudonym.
The collection process of the tweets was conducted as follows.
We retrieved archived tweets~\citep{Wu2020} from \url{www.polititweet.com}, a website dedicated to archiving Twitter postings of public figures. 
These archived tweets may be truncated due to the API settings used by polititweet.
As such, we re-obtained the non-truncated tweets (where still available) from the Twitter API using the tweet ID from the polititweet postings.

\subsection{\pubfig Subset Selection}
\label{subsec:subset-selection}

Labeling more than $300,000$ postings is time and cost-prohibitive; therefore, we label a subset -- dubbed \pubfig.
Hate speech is rare, and hate speech from public figures is even rarer; 
hence, selecting the subset through random sampling would yield very few hateful tweets.
To build a more balanced set, we employ an active sampling procedure.
First, we train a binary classifier specialized for classifying unseen datasets using our MTL framework (see the MTL-NCH classifier in \cref{subsec:classification_on_unseen_data} for further details).
We train this classifier using the eight hate speech datasets described in \cref{sec:existing_datasets} for 10 epochs at a learning rate of $2e-5$ and a batch size of $512$, as described in \cref{subsec:training_pipeline}. 
We refer to this classifier as \textit{MTL-NCH[8]}. 

We apply MTL-NCH[8] to \pubfigfull, machine-labeling it with the ``harmless'' and ``problematic'' labels.
We build a balanced subset as follows.
First, we build the likely positive set by selecting all tweets labeled as problematic.
Second, we build the likely negative tweet set by undersampling an equal number of tweets labeled as harmless for each figure so that the dataset is not overly skewed towards any single figure. 
We obtain a final subset of $20,327$ tweets.
Next, we human-annotate this subset via crowd-sourcing using Amazon Mechanical Turk.

\subsection{\pubfig Amazon Mechanical Turk Labeling} 
\label{subsec:mturk_labelling}
We preprocess the tweets to remove links, identification data, and non-textual data such as videos and images. 
Links were substituted with a replacement token to indicate where a link existed in the text.
Workers were set to label each tweet as either ``Hate'', ``Abuse'', or ``Neutral''.
We provided the workers with definitions of each class and positive and negative examples (we show the instructions and worker interface in the supplementary materials).
The definitions were given as follows:
\textit{``Hate"}: content that directly or indirectly attacks, discriminates, incites violence and/or hate against a person or group on the basis of who they are.
\textit{``Abuse"}: content that is abusive but not hateful based on the criteria above. 
Any content that bullies, harasses, insults, or is otherwise offensive to an individual or group but not on the basis of their identity. 
\textit{``Neutral"}: any content not fitting the other two classes.
We restrict participation to workers in English-speaking countries (the UK, the United States, Canada, and Australia) to increase the likelihood of native speakers with socio-political background knowledge.
We further require workers to have a $98\%$ approval rate and have completed at least $5000$ MTurk HITS.
In total, $456$ workers contributed to the labeling process, and $8$ workers labeled each tweet. 
Further details regarding the MTurk Labeling process, such as payment and task batch size, can be found in the supplementary materials.
We determined the final annotations based on a two-stage majority vote procedure described in the next section.

\subsubsection{Tie-Breaking}
Hate speech identification is a difficult problem even for humans; therefore, it is expected to have diverse labels among the 8 annotators of each tweet.
A simple majority vote among the annotators is unlikely to yield the expected results.
For example, a tweet labeled by the workers as 3:2:3 (neutral : abuse : hate) is unlikely to be neutral, as five annotators considered it abusive or harmful.
We start from the hypothesis that it is easier to distinguish between neutral and problematic speech -- we dub speech as \emph{problematic} when it is either abusive or hateful --, than between abuse and hate.
As such, we devise a two-stage procedure that we dub \emph{tie-breaking}.
First, we break between neutral and problematic, then between hateful and abusive.
In stage 1, we compare the number of annotators that selected neutral versus those that selected abuse or neutral and select the higher number.
If we selected abuse/hate, then we compare the amount of abuse versus the amount of hate.
In the example above, we chose problematic in stage one, and hate in stage two.

However, at any stage, there may be an equal amount of annotators for both options.
As a result, we explored several tie-breaking strategies:
\emph{NMV-H}: breaking na\"ively through a single majority vote between all 3 classes to more hateful class (e.g., 4:4:2 yields abuse);
\emph{NMV-L}: breaking na\"ively through a single majority vote between all 3 classes to less hateful class (e.g., 4:4:2 yields neutral);
\emph{HH}: breaking to the more hateful class for both stages -- in case of equality, select problematic for stage 1, and hate at stage 2 (e.g., 3:2:3 yields hate);
\emph{LL}: breaking to the less hateful class for both stages -- neutral at stage 1, and abuse at stage 2 (3:2:3 yields neutral);
\emph{HL}: breaking first to the more hateful class at stage 1 (i.e., problematic), and to the less hateful class at stage 2 (i.e., abuse) (3:3:3 would yield abuse); and;
\emph{LH}: breaking first to the less hateful class at stage 1 and to the more hateful class at stage 2 (3:3:3 would yield neutral).

To select a tie-breaking strategy, we took a small sample of 200 instances from the \davidson dataset and re-labeled them following our methodology described in \cref{subsec:mturk_labelling} using the Amazon MTurkers.
The intuition is that the best tie-breaking strategy will yield the closest results to the \davidson annotations.
We chose the \davidson dataset due to its closeness to our labeling task. 
Both the \davidson and our labeling tasks use tweets, labeled with 3 classes: neutral, offensive, hate for \davidson and neutral, abusive, hate for our task. 
The offensive class was mapped to the abusive class in the labeling process, as both classes followed roughly similar definitions. 
We treat each tie-breaking strategy as a classifier and the already assigned labels from the \davidson dataset as the ground truth.
We compute typical classification error metrics (such as the F1-score), and consider that the tie-breaking method with the highest F1 is the best suited.
\cref{tab:tie_breaking} shows that HL achieves the highest three-class F1-score -- if a post has an equal number of neutral and abuse/hate, we chose abuse/hate in stage 1; in stage 2, if it has an equal number of abuse and hate annotations, we chose abuse.
We use this strategy to build the final annotation for each post in \pubfig.

\begin{table}[]
	\centering
	\caption{
		Comparison of tie-breaking strategies applied to a sample of the \davidson dataset for the 3-class problem (``neutral", ``abusive", and ``hate") and the binary problem (``neutral' vs ``abusive''/``hate"). 
		Highest scores shown in boldface.
	}
	\label{tab:tie_breaking}
		
	\begin{tabular}{lrrrrrrr}
		\toprule
		\multicolumn{1}{c}{Strategy} & \multicolumn{1}{c}{NMV-H} & \multicolumn{1}{c}{NMV-L} & \multicolumn{1}{c}{HH} & \multicolumn{1}{c}{HL} & \multicolumn{1}{c}{LH} & \multicolumn{1}{c}{LL} \\
		\midrule
		3 Class F1       &   $0.389$ & $0.431$           & $0.468$                 & $\textbf{0.481}$                 & $0.453$                 & $0.422$                 \\
		Binary F1        &   $0.568$ & $0.674$           & $0.709$                 & $\textbf{0.709}$                 & $0.615$                 & $0.615$           \\     
		\bottomrule 
	\end{tabular}
\end{table}
 
\begin{table}[h]
	\caption{
		Comparison of mean agreement with majority vote for MTurk worker annotations before and after blocklisting underperforming workers.
		The columns show the mean agreement with the majority when using 3 labels (``neutral", ``abusive", ``hate") and two labels (``neutral'' vs ``abusive''/``hate").
	}
	\label{tab:agreement_compare}
	\centering
	\begin{tabular}{lrrr}
		\toprule
		& \renewcommand\cellalign{cc}\makecell{3 Class\\Mean Agreement} & \renewcommand\cellalign{cc}\makecell{Binary\\Mean Agreement} & \renewcommand\cellalign{cc}\makecell{Krippendorff's\\Alpha} \\
		\midrule
		With underperforming workers & $0.649$        & $0.701$   &  $0.063$ \\
		After blocklisting and re-labeling  & $0.803$        & $0.813$   &  $0.124$   \\
		\midrule
		Improvement & $23.72\%$ & $15.97\%$ & $96.82\%$\\
		\bottomrule
	\end{tabular}
\end{table} 

\begin{figure}[t]
	\centering	
	\newcommand\myheight{0.195}
	\subfloat[]{
		\includegraphics[height=\myheight\textheight]{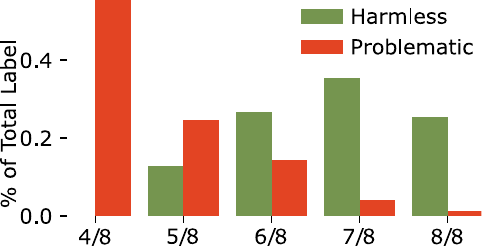}%
		\label{fig:consensus_binary}%
	}%
	\subfloat[]{
		\includegraphics[height=\myheight\textheight]{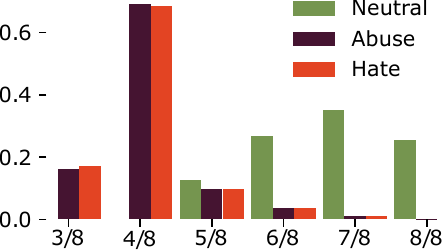}%
		\label{fig:consensus_3_class}%
	}%
	\caption{
		\textbf{Amazon Turkers' mean agreement with the majority for the final labels in the \pubfig dataset.}
		Eight workers label each tweet via majority vote. 
		The x-axis shows how many workers selected the majority label.
		The y-axis shows the percentage of total instances assigned to a label with a given majority vote split.
		\textbf{(a)} binary: harmless (neutral) vs. problematic (abuse and hate); 
		\textbf{(b)} 3 class labeling (neutral, abuse, hate).
	}
	\label{fig:consensus}	
\end{figure}

\subsubsection{Filtering Underperforming Workers} 
The nature of Amazon Mechanical Turk incentivizes workers to complete work as quickly as possible to maximize gains. 
As such, some workers will take shortcuts and deliver low-quality work. 
The most common examples of low-quality work were randomly selecting among the three classes or arbitrarily between 2 classes (alternating the class selection between subsequent answers).
We found that increasing worker payment did not affect work quality and resorted to blocklisting underperforming workers.
We performed a full annotation round (with no blocklisting) and identified underperforming users as described below.
After the blocklisting, we had to reannotate some of the exemplars.

The annotation task was done in batches of 50 to 100 HITs (Human Intelligence Tasks), with each HIT consisting of 10 tweets for labeling.
For each batch, we examined the top 25 workers who completed the most work and blocklisted them if the work was poor quality.
We identified suspect workers using two metrics:
Firstly, we examined the workers' Krippendorff's Alpha \citep{krippendorff2011computing} -- their mean consensus with the final instance label of the majority vote. 
We operate under the assumption that most workers do not behave maliciously; hence, users with a low Krippendorff's Alpha value are suspicious as it indicates that they tend to systematically disagree with the majority of other workers across many tweets. 
Secondly, we examined the distribution of user-assigned labels. 
We know that hate speech is rare among public figures; therefore, we consider as suspicious any worker with an unusual class distribution;
for example, skewed towards abuse or hate, equal among the three classes, or constantly labeling only one or two classes.
We manually examined the work of suspected workers for quality. 
In particular, we examined posts labeled as hateful and abusive for the presence of overtly benign content. 
Overtly underperforming workers were blocklisted and unable to further work on our task.
Tweets labeled by more than two underperforming workers were redone. 
To analyze the effectiveness of this filtering and reannotation process, we examined the mean agreement with the majority vote across all instances for both the 3 class problem and the binary problem of neutral vs non-neutral. 
A higher agreement indicates a more robust label since we assume that most workers do not behave maliciously.
\cref{tab:agreement_compare} shows that blocklisting underperforming workers improves the mean agreement with the majority by $23.72\%$ and the inter-worker Krippendorff's Alpha by $98.82\%$.

\subsection{\pubfig Dataset Profiling.}
\label{subsec:pubfigs_profiling}
The crowdsourced annotation yielded $17,963$ neutral instances, $1,422$ abusive instances, and $942$ hate instances.
We find that all figures had consistently more neutral than problematic (abuse, hate) labeled tweets -- highlighting the class imbalance.
The vast majority of problematic tweets ($96.62\%$) originate from right-leaning figures: $915$ (out of $942$) hateful tweets and $1369$ ($1422$) abusive tweets.
Within the hate class, every right-leaning figure in the dataset had at least $10$ hateful tweets. 
AC, a controversial right-leaning political commentator, accounts for just under half of all hate-labeled instances ($464$ out of $942$).
Comparatively, the left-leaning figure with the most hate tweets was IO ($23$), although these are most likely false positives (see \cref{subsec:real_world_classification}).
Four left-leaning figures (BR, TS, BO, and MO) had no tweets labeled hate. 

\textbf{Hate Speech Annotation -- A Difficult Problem.}
Annotating hate speech is a challenging task. 
This difficulty can be quantified by examining the inter-annotator agreement, as a difficult task often leads to lower consensus among workers. 
We use the mean agreement with the majority vote over all tweets as a measure of difficulty rather than the more widely used Krippendorff's Alpha.
The latter is skewed to low values when there is a low overlap of annotated instances between workers. 

We analyze workers' agreement in two tasks. 
The first is differentiating between harmless content (neutral) and problematic content (abuse and hate). 
The second task is differentiating between the three classes. 
We expect workers to perform better in the first task as it is easier to identify harmless from problematic content than differentiating between flavors of problematic (hate and abuse). 
\cref{fig:consensus} shows the average worker agreement with the majority vote for the first binary task (\cref{fig:consensus_binary}) and the second three-class task (\cref{fig:consensus_3_class}). 
The x-axis shows the number of annotators who agreed with the label assigned via the majority vote; 
the y-axis shows the percentage of instances with the given agreement between annotators. 

Intuitively, a difficult decision is signaled by a high percentage of instances (high y-axis) for which few annotators agree (low x-axis).
\cref{fig:consensus_binary} shows that annotators have a higher agreement for the harmless class (most tweets achieve a consensus of seven annotators out of eight, $7/8$) than for the problematic class (consensus of $4/8$).
We posit that the language used by public figures may not be overtly abusive or hateful, making it difficult for annotators to identify.
In the three-class task (\cref{fig:consensus_3_class}), the abuse and hate classes have similar distributions over the final majority vote, indicating that both classes are similarly challenging to identify.
Finally, about $18\%$ of the abuse and hate instances have a low agreement outcome ($3/8$), which can occur when as many as six annotators out of eight agree that the instance is problematic.
This suggests that identifying the explicit facet of problematic content is difficult even when there is overall agreement that the tweet is problematic.

\subsection{Machine-labeling \pubfigfull.}
\label{subsec:full-pubfig}
The final action in \cref{fig:pubfigs_construction_diagram} is machine-labeling the entire \pubfigfull dataset.
To achieve this, we train an MTL classifier (see \cref{sec:methodology}) using the eight public datasets introduced in \cref{sec:existing_datasets} and the human-labeled \pubfig dataset.
We optimize the MTL model for each of the 9 datasets simultaneously; once the training is complete, we leverage the classification head corresponding to \pubfig to 
machine-label all the $305,235$ tweets in the \pubfigfull dataset according to the \pubfig labels.
We obtain $298,803$ neutral, $5,299$ abusive, and $1,133$ hateful tweets. 
The minority nature of the abusive and hateful tweets -- comprising $1.74\%$ and less than $1\%$ of the total tweets, respectively -- was as anticipated given Twitter's policy on hateful conduct and its efforts to eliminate hate speech from its platform \citep{twitter_hate_policy}.

The \pubfigfull dataset is about 15 times larger than the \pubfig subset; 
however, it contains only marginally more hate tweets ($1,133$ in \pubfigfull vs $942$ in \pubfig) and 4 times more abusive tweets ($5,299$ in \pubfigfull vs $1,422$ in \pubfig).
The classifier labeled the vast majority of tweets as neutral.
This is expected given the construction of the \pubfig subset.
In \cref{subsec:subset-selection}, we used an MTL classifier trained on eight public datasets to identify potentially problematic speech.
As it turns out, that classifier had a good performance, missing only very few hateful tweets.
We perform additional error analysis in \cref{subsec:unseen_data_classification}.
\begin{table*}[tbp]
    \caption{
        \textbf{Unseen dataset performance of existing works and our MTL pipeline.}
        MTL is trained on all datasets except the testing dataset.
        The best results are in bold; 
        we report the same number of decimals as the original work. 
        For MTL we show $\pm\textit{st.dev}$.
        We could not obtain the \textsc{Founta} and \textsc{OLID} datasets, and we do not include them in our work.
        The number of wins in parentheses.}
    \label{tab:unseen_vs_sota}
    \small
    \setlength{\tabcolsep}{3pt}
    \begin{tabular}{lllllll}
        \toprule
        \textbf{Prior Work}                   & \textbf{Model} & \thead[tl]{\textbf{Train}\\\textbf{Dataset}} & \thead[tl]{\textbf{Test}\\\textbf{Dataset}} & \textbf{Metric} & \thead[tl]{\textbf{Reported}\\($3/8$)} & \thead[tl]{\textbf{MTL-NCH}\\($5/8$)} \\
        \midrule
        \citet{10.1145/3331184.3331262}  & \makecell[tl]{\citet{Badjatiya2017}\\LSTM + GBDT (binary)} & \makecell[tl]{\waseem\\+~\davidson}    & \hateval              & Macro F1        & 0.516                     & \textbf{0.645} $\pm\textit{0.006}$          \\
        \citet{10.1145/3331184.3331262}  & \makecell[tl]{\citet{Agrawal2018}\\Bi-LSTM baseline (binary)}      & \makecell[tl]{\waseem\\+~\davidson}    & \hateval              & Macro F1        & 0.541                     & \textbf{0.645} $\pm\textit{0.006}$          \\
        \citet{swamy-etal-2019-studying} & BERT binary                                         & \waseem                & \davidson             & Macro F1        & 0.5296                    & \textbf{0.6822} $\pm\textit{0.0197}$         \\
        \citet{swamy-etal-2019-studying} & BERT binary                                         & \textsc{Founta}        & \davidson             & Macro F1        & 0.5824                    & \textbf{0.6822} $\pm\textit{0.0197}$         \\
        \citet{swamy-etal-2019-studying} & BERT binary                                         & \textsc{OLID}          & \davidson             & Macro F1        & 0.5982                    & \textbf{0.6822} $\pm\textit{0.0197}$         \\
        \citet{swamy-etal-2019-studying} & BERT binary                                         & \davidson              & \waseem               & Macro F1        & \textbf{0.6928}           & 0.4995 $\pm\textit{0.0085}$                  \\
        \citet{swamy-etal-2019-studying} & BERT binary                                         & \textsc{Founta}        & \waseem               & Macro F1        & \textbf{0.6049}           & 0.4995 $\pm\textit{0.0085}$                  \\
        \citet{swamy-etal-2019-studying} & BERT binary                                         & \textsc{OLID}          & \waseem               & Macro F1        & \textbf{0.6269}           & 0.4995 $\pm\textit{0.0085}$                  \\
        \bottomrule
    \end{tabular}
\end{table*} 
\section{Experiments and Results}
\label{sec:experiments_and_results}
This section discusses the evaluation of our MTL approach.
First, in \cref{subsec:unseen_data_classification}, we present the hate speech detection on unseen datasets and the improvements in train-test splits.
Next, in \cref{subsec:real_world_classification}, we apply the MTL model on the \pubfigfull dataset and analyze the inappropriate speech patterns of a sample of American public figures. All reported results are mean averages over $10$ experimental runs. 

\textbf{Metrics.}
We use \textit{macro averaged F1} as our the evaluation metric, defined for $N$ classes as:
\begin{equation*}
    \text{Macro F1} = \frac{1}{N} \sum_{i=1}^{N} F1_i = \frac{1}{N} \sum_{i=1}^{N} \frac{2 \times precision_{i} \times recall_i}{precision_{i} + recall_i} \enspace
\end{equation*}

The F1 score is the harmonic mean between Precision and Recall and can be understood as a trade-off between the two metrics.
Macro averaging is selected over other averaging methods as it assigns equal importance to each class irrespective of their size. 
This is beneficial in our evaluation tasks as hate speech datasets are known to often be highly imbalanced \citep{Yuan2023,madukwe-etal-2020-data}, with only a few positive examples for some classes.

\subsection{Unseen Dataset Classification.}
\label{subsec:unseen_data_classification}

\begin{figure}[th]
	\centering	
	\includegraphics[scale=0.825]{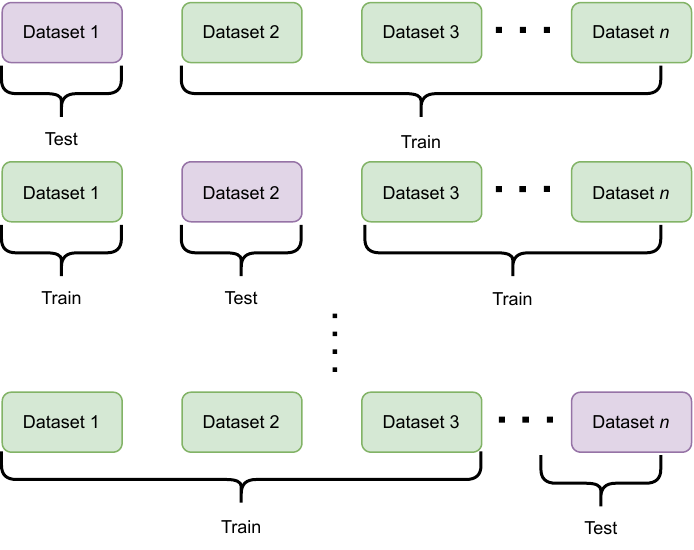}	
	\caption{
		The problem setup for unseen dataset classification. We adopt a leave-one-out evaluation scheme for evaluating the pipeline's ability to generalize on a completely new dataset.}
	\label{fig:kfold_experiment}
\end{figure}

\begin{table*}[tbp]
    \caption{
        \textbf{Macro-F1 prediction performances on a target dataset, unseen during training} (shown by column headers).
        \textit{(above horizontal ruler)} Our two MTL flavors (NCH and MV) trained on all datasets except the target dataset.
        \textit{(below horizontal ruler)} Transferability between pairs of datasets.
        A single dataset baseline (see \cref{subsec:training_pipeline}) is trained on the source dataset (rows) and tested on the target dataset (columns).
        The best results are in bold.
    }
    \label{tab:unseen_vs_baseline}
    \small
    \setlength{\tabcolsep}{4pt}
    \begin{tabular}{clrrrrrrrrrr}
        \toprule
         &                      & \multicolumn{9}{c}{\footnotesize Testing Dataset}  \\
         & \thead[tc]{\textbf{Model}}       & \thead[tc]{\textbf{\davidson}}                         & \thead[tc]{\textbf{\waseem}}                        & \thead[tc]{\textbf{\reddit}}                        & \thead[tc]{\textbf{\gab}}                         & \thead[tc]{\textbf{\fox}}                            & \thead[tc]{\textbf{\textsc{Storm-}}\\\textbf{\textsc{Front}}}                    & \thead[tc]{\textbf{\mandl}}                         & \thead[tc]{\textbf{\textsc{Hat-}}\\\textbf{\textsc{Eval}}}                       & \thead[tl]{\textbf{\textsc{Pub-}}\\\textbf{\textsc{Figs-L}}}                        & \thead[tc]{\textbf{Wins}} \\
        \midrule
        \multirow{2}*{\footnotesize\rotatebox{90}{MTL}}
         & \textbf{MTL-NCH}     & \textbf{\cellcolor[HTML]{C1DA81}0.6822}    & \cellcolor[HTML]{F98470}0.3801          & \textbf{\cellcolor[HTML]{71C27C}0.8456} & \textbf{\cellcolor[HTML]{63BE7B}0.8738} & \textbf{\cellcolor[HTML]{E2E383}0.6150} & \textbf{\cellcolor[HTML]{C1DA81}0.6826} & \cellcolor[HTML]{FEDC81}0.5312          & \textbf{\cellcolor[HTML]{D4DF82}0.6449} & \cellcolor[HTML]{E1E383}0.6175          & 6                \\
         & \textbf{MTL-MV}      & \cellcolor[HTML]{D3DF82}0.6455             & \cellcolor[HTML]{FA9273}0.4048          & \cellcolor[HTML]{7BC57D}0.8263          & \cellcolor[HTML]{67C07C}0.8660          & \cellcolor[HTML]{E8E583}0.6030           & \cellcolor[HTML]{C4DA81}0.6771          & \cellcolor[HTML]{FCC07B}0.4834          & \cellcolor[HTML]{DAE182}0.6315          & \textbf{\cellcolor[HTML]{DEE283}0.6231} & 1                \\
        \midrule
        \multirow{9}*{\footnotesize\rotatebox{90}{ \begin{tabular}[x]{@{}c@{}}BERT Baseline \\Trained On:\end{tabular}}}
         & \textbf{\davidson}   &                                            & \cellcolor[HTML]{FEEA83}0.5556          & \cellcolor[HTML]{EEE683}0.5914          & \cellcolor[HTML]{C6DB81}0.6731          & \cellcolor[HTML]{FDC67C}0.4932           & \cellcolor[HTML]{FBB279}0.4597          & \cellcolor[HTML]{F9EA84}0.5690          & \cellcolor[HTML]{FEE282}0.5414          & \cellcolor[HTML]{FEE583}0.5469          & 0                \\
         & \textbf{\waseem}     & \cellcolor[HTML]{E3E383}0.6136             &                                         & \cellcolor[HTML]{EAE583}0.6000          & \cellcolor[HTML]{D5DF82}0.6427          & \cellcolor[HTML]{FEE883}0.5519           & \cellcolor[HTML]{FEDF81}0.5356          & \cellcolor[HTML]{FDD07E}0.5099          & \cellcolor[HTML]{F4E884}0.5784          & \cellcolor[HTML]{FDEB84}0.5611          & 0                \\
         & \textbf{\reddit}     & \cellcolor[HTML]{E3E383}0.6135             & \cellcolor[HTML]{FDC77D}0.4957          &                                         & \cellcolor[HTML]{84C87D}0.8083          & \cellcolor[HTML]{FDD780}0.5229           & \cellcolor[HTML]{FFEB84}0.5559          & \cellcolor[HTML]{FCC47C}0.4900          & \cellcolor[HTML]{F7E984}0.5741          & \cellcolor[HTML]{FEE182}0.5402          & 0                \\
         & \textbf{\gab}        & \cellcolor[HTML]{F8E984}0.5720             & \cellcolor[HTML]{FBB279}0.4595          & \cellcolor[HTML]{75C47D}0.8375          &                                         & \cellcolor[HTML]{FDCE7E}0.5075           & \cellcolor[HTML]{FBEA84}0.5645          & \cellcolor[HTML]{FAA075}0.4277          & \cellcolor[HTML]{FAEA84}0.5664          & \cellcolor[HTML]{FDD57F}0.5185          & 0                \\
         & \textbf{\fox}        & \cellcolor[HTML]{FAA075}0.4285             & \cellcolor[HTML]{FA9E75}0.4249          & \cellcolor[HTML]{FA9D75}0.4234          & \cellcolor[HTML]{FCB579}0.4651          &                                          & \cellcolor[HTML]{F98770}0.3865          & \cellcolor[HTML]{FA9974}0.4159          & \cellcolor[HTML]{FBAC77}0.4490          & \cellcolor[HTML]{F98B71}0.3926          & 0                \\
         & \textbf{\stormfront} & \cellcolor[HTML]{FBAF78}0.4533             & \cellcolor[HTML]{FEE582}0.5467          & \cellcolor[HTML]{F3E884}0.5822          & \cellcolor[HTML]{D2DE82}0.6487          & \cellcolor[HTML]{F7E984}0.5740           &                                         & \cellcolor[HTML]{FDD07E}0.5104          & \cellcolor[HTML]{FAEA84}0.5664          & \cellcolor[HTML]{FBEA84}0.5659          & 0                \\
         & \textbf{\mandl}      & \cellcolor[HTML]{F8696B}0.3336             & \cellcolor[HTML]{FCBF7B}0.4822          & \cellcolor[HTML]{FA9373}0.4066          & \cellcolor[HTML]{FBB179}0.4582          & \cellcolor[HTML]{FA9072}0.4010           & \cellcolor[HTML]{F8736D}0.3518          &                                         & \cellcolor[HTML]{FBAF78}0.4546          & \cellcolor[HTML]{F87A6E}0.3633          & 0                \\
         & \textbf{\hateval}    & \cellcolor[HTML]{F1E784}0.5849             & \cellcolor[HTML]{F2E884}0.5824          & \cellcolor[HTML]{F9E984}0.5700          & \cellcolor[HTML]{F4E884}0.5796          & \cellcolor[HTML]{FEE983}0.5532           & \cellcolor[HTML]{FEE582}0.5466          & \cellcolor[HTML]{FEDE81}0.5348          &                                         & \cellcolor[HTML]{FEE382}0.5432          & 0                \\
         & \textbf{\pubfig}     & \cellcolor[HTML]{D9E082}0.6351             & \textbf{\cellcolor[HTML]{E7E583}0.6048} & \cellcolor[HTML]{EBE683}0.5970          & \cellcolor[HTML]{CCDD82}0.6600          & \cellcolor[HTML]{FEEA83}0.5546           & \cellcolor[HTML]{FED880}0.5249          & \textbf{\cellcolor[HTML]{ECE683}0.5963} & \cellcolor[HTML]{F1E784}0.5858          &                                         & 2                \\
        \bottomrule

    \end{tabular}
\end{table*} %

\begin{table*}[t]
    \caption{
        \textbf{MTL against 9 baselines (column ``Model in related work'') for targeted hate speech prediction} (see the ``Setup'' in \cref{subsec:improvement_of_target_task}).
        We copy the performances reported by the works (column ``Reported'').
        We report using the performance metrics and the same number of decimals as reported in the original papers (column ``Metric'') $\pm \textit{standard deviation}$.
        The number of wins is shown in parentheses.
        \citet{10.1145/3331184.3331262} use a binary hate/non-hate label mapping that we replicate for the corresponding MTL comparisons. 
        \citet{Waseem2018} do not specify the averaging method; hence we assume \emph{micro-F1}.
    }
    \label{tab:targeted_vs_sota}
    \small
    \setlength{\tabcolsep}{3pt}
    \begin{tabular}{llllll}
        \toprule
        \textbf{Work}           & \textbf{Model in Related Work}                                            & \textbf{Dataset} & \textbf{Metric} & \thead[tl]{\textbf{Reported}\\($7/20$)} & \textbf{MTL} ($12/20$) \\
        \midrule
        \citet{mozafari_bert-based_2019} & BERT+CNN       & \waseem           & Weighted F1     & \textbf{0.88}       & 0.83 $\pm\textit{0.01}$          \\
        \citet{mozafari_bert-based_2019} & BERT+CNN       & \davidson         & Weighted F1     & \textbf{0.92}       & 0.90 $\pm\textit{0.01}$        \\
        \citet{MacAvaney2019}   & BERT finetune           & \hateval        & Macro F1        & 0.7452   & \textbf{0.7526} $\pm\textit{0.0154}$         \\
        \citet{MacAvaney2019}   & mSVM                    & \hateval        & Macro F1        & 0.7481   & \textbf{0.7526} $\pm\textit{0.0154}$         \\
        \citet{10.1007/978-3-319-93417-4_48} & CNN+sCNN   & \waseem           & Micro F1        & 0.83      & 0.83 $\pm\textit{0.01}$           \\
        \citet{10.1007/978-3-319-93417-4_48} & CNN+sCNN   & \waseem           & Macro F1        & 0.77              & \textbf{0.78} $\pm\textit{0.01}$  \\
        \citet{10.1007/978-3-319-93417-4_48} & CNN+sCNN   & \davidson         & Micro F1        & \textbf{0.94}     & 0.90   $\pm\textit{0.01}$         \\
        \citet{10.1007/978-3-319-93417-4_48} & CNN+sCNN   & \davidson         & Macro F1        & 0.64              & \textbf{0.74} $\pm\textit{0.01}$  \\
        \citet{10.1145/3331184.3331262} & \makecell[tl]{\citet{Badjatiya2017}\\LSTM + GBDT baseline (binary)}      & \waseem           & Micro F1        & 0.807             & \textbf{0.828} $\pm\textit{0.008}$ \\
        \citet{10.1145/3331184.3331262} & \makecell[tl]{\citet{Badjatiya2017}\\LSTM + GBDT baseline (binary)}      & \waseem           & Macro F1        & 0.731             & \textbf{0.802}  $\pm\textit{0.010}$ \\
        \citet{10.1145/3331184.3331262} & \makecell[tl]{\citet{Agrawal2018}\\Bi-LSTM baseline (binary)} & \waseem   & Micro F1        & \textbf{0.843}    & 0.828  $\pm\textit{0.008}$         \\
        \citet{10.1145/3331184.3331262} & \makecell[tl]{\citet{Agrawal2018}\\Bi-LSTM baseline (binary)} & \waseem   & Macro F1        & 0.796    & \textbf{0.802}  $\pm\textit{0.010}$          \\
        \citet{Madukwe_GA_BERT}         & GA (BERT+CNN+LSTM)                                       & \davidson         & Weighted F1     & 0.87              & \textbf{0.90} $\pm\textit{0.01}$ \\
        \citet{Madukwe_GA_BERT}         & GA (BERT+CNN+LSTM)                                       & \davidson         & Macro F1        & 0.73              & \textbf{0.74} $\pm\textit{0.01}$  \\
        \citet{Waseem2018} & BOW        & \waseem           & Micro F1    & \textbf{0.87}     & 0.83 $\pm\textit{0.01}$          \\
        \citet{Waseem2018} & BOW        & \davidson         & Micro F1    & 0.89              & \textbf{0.90} $\pm\textit{0.01}$   \\
        \citet{Yuan2023} & Bi-LSTM   & \waseem           & Macro F1    & 0.7809            & \textbf{0.7823} $\pm\textit{0.0133}$          \\
        \citet{Yuan2023} & Bi-LSTM   & \davidson         & Macro F1    & 0.7264             & \textbf{0.7450} $\pm\textit{0.0052}$   \\
        \citet{KAPIL2020106458} & SP-MTL + CNN & \waseem           & Macro F1    & \textbf{0.8916}            & 0.7823 $\pm\textit{0.0133}$          \\
        \citet{KAPIL2020106458} & SP-MTL + CNN & \davidson         & Macro F1    & \textbf{0.9115}             & 0.7450 $\pm\textit{0.0052}$   \\

        \bottomrule
    \end{tabular}
\end{table*} 
Here, we examine whether the MTL pipeline can train models that detect hateful and abusive speech in previously unseen datasets. 

\textbf{Experiment Setup and Evaluation.}
We use a leave-one-out setup:
given $n$ datasets, we train the shared BERT representation using $n-1$ datasets, leaving out the dataset $d_t$. 
We use the MTL-trained BERT -- in the NCH or MV classification scheme (see \cref{subsec:classification_on_unseen_data}) -- to evaluate on $d_t$.
Note that no portion of $d_t$ is observed during training; thus, $d_i$ is an entirely new dataset.
We rotate the left-out dataset $d_t$ until we have evaluated all datasets. 
A diagram of the evaluation scheme can be seen in \cref{fig:kfold_experiment}.
We also evaluate the transferability between datasets by training single dataset baselines (see \cref{subsec:training_pipeline}).
We train on one dataset and test on another for each possible pair of datasets ($81$ unique pairs).
Datasets with multiple problematic classes are binarized as problematic/harmless using the mapping in \cref{tab:datasets}.

\textbf{Outperforming the State-of-the-Art.}
We start by assessing the performance of MTL against existing state-of-the-art that examined predicting hate speech on unseen datasets.
In \cref{tab:unseen_vs_sota}, we select prior works that evaluate using the same datasets used in our work.
Our model outperforms the existing models in 5 out of 8 comparisons, achieving nearly $10\%$ higher performance on \hateval and \davidson datasets. 
We underperform solely on the \waseem dataset, which is known to have particular labeling and contains many false positives \citep{Yuan2023,Waseem-2016-racist}.
Notably, the \waseem dataset was not constructed for hate speech classification but rather to assess the agreement between amateur and expert annotators, as discussed in \citep{Yuan2023} and the supplementary appendix. 
We contend that a generalized model is not ideal for such datasets, leading to lower prediction performance.

In \cref{tab:unseen_vs_baseline}, we further investigate the classification performances of MTL and single dataset baselines. 
We note two key observations. 
Firstly, the MTL flavors consistently outperform the single dataset baselines, achieving the best results on seven out of nine datasets. 
This finding indicates that MTL successfully enhances the predictive ability on unseen datasets. 
For the remaining two datasets (\waseem and \mandl), the single dataset baseline trained on \pubfig, which is introduced in this work, achieves the best performance. 
Secondly, we find that MTL-NCH outperforms MTL-MV in seven out of the nine datasets, and it is within $2\%$ of the MTL-MV performance on the remaining two (\waseem and \pubfig). 
As a result, we focus solely on the MTL-NCH flavor in all discussions in this paper.

\textbf{Error Analysis on \pubfig.}
We compare the classification results of the MTL-NCH classifier against the human annotated labels in \pubfig to gain insight into the types of errors that the classifier makes. This classifier is equivilant to MTL-NCH[8] described in \cref{subsec:subset-selection}.
\cref{tab:pubfigs_l_confusion_table} shows the confusion matrix of MTL-NCH[8].
The classifier shows good performance on both hateful and abusive tweets, with the classifier correctly identifying $82\%$ of abusive or hateful tweets as problematic.
We find that the main error was a false positive by MTL-NCH where it considered a human labeled neutral tweet to be problematic.
Of the 10170 tweets classified as problematic, the vast majority of tweets (8230/1170) were deemed to be not hateful or abusive by the human annotators. 
This indicates that the MTL-NCH classifier is overly liberal with assigning the problematic classification in the \pubfig dataset.

\subsection{Generalization to Unseen Datasets.}

\begin{figure}[t]
	\centering	
	\includegraphics[width=0.9\columnwidth]{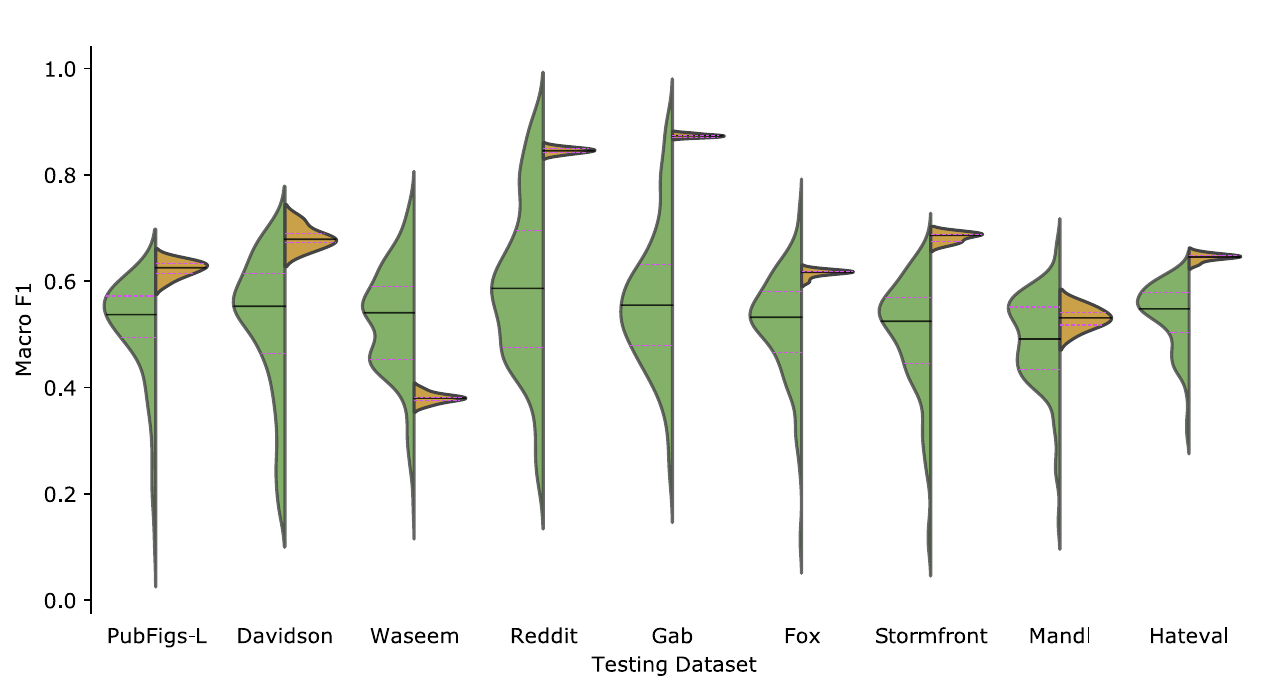}	
	\caption{
		\textbf{Unseen classification performance} (macro-F1) on each target dataset by the single dataset trained baselines (green, left) and the MTL pipeline (orange, right). 
		The magenta lines indicate quartiles.
	}
	\label{fig:unseen_violin}
\end{figure}

We further study how transferable each dataset is to each other. Given that no definition of hate is universally agreed upon, each dataset brings with it its own biases. We hypothesize that certain datasets may generalize better to others due to shared biases in their definitions.
\cref{fig:unseen_violin} presents the distribution of macro-F1 scores using violin plots, with each violin representing a target dataset how transferable each target dataset is from other datasets.
The left half of each violin illustrates the performance distribution of all single dataset baselines, whereas the right half shows the performance of MTL-NCH. 
We observe that MTL-NCH exhibits higher mean prediction performance and lower variance than single baselines. 
Additionally, the single baselines for \waseem and \mandl exhibit two visible modes, while \hateval, \reddit, and \gab show less pronounced bimodality. 
This phenomenon arises from the uneven transferability of models -- some datasets generalize better to specific others, potentially due to similarities in their hate speech facets or annotation procedures.
This is shown in \cref{tab:unseen_vs_baseline}, where higher performance indicates better generalization to unseen datasets. 
For instance, \gab and \reddit exhibit bidirectional generalization, achieving over $80\%$ macro-F1 performance in both directions.
This is unsurprising, given that they were proposed in the same work. 
Other dataset pairs generalize only one way, e.g., \davidson to \gab, \pubfig to \davidson, \pubfig to \gab, and \waseem to \reddit.
This may occur when one dataset covers more hate speech facets than the other.
We also analyze dataset similarity in terms of their vocabulary usage.
We measure the Ruzicka similarity~\citep{ruzicka1958anwendung}, a weighted Jaccard similarity quantifying the overlap in the terms occurring in the problematic and harmless classes.
While \gab and \reddit have similar language use for both classes, \gab and \pubfig are only similar for the harmless class (Ruzicka$=0.37$ for harmless, $0.08$ for problematic), which still leads to a moderate generalization (macro-F1$=0.66$). 
More details on the Ruzicka similarities are available in the supplementary appendix.

\begin{table*}[t]
	\centering
	\caption{MTL-NCH[8] label counts compared to the final MTurk human assigned labels. Red italics are errors.}
	\label{tab:pubfigs_l_confusion_table}
	\begin{tabular}{llll}
		\toprule
		\multirowcell{2}{MTL-NCH[8]\\ Classification} & \multicolumn{3}{c}{MTurk Assigned Label} \\
		& \textbf{Neutral} & \textbf{Abuse} & \textbf{Hate} \\
		\midrule
		\textbf{Neutral}     & 9733 & \textit{\textcolor{red}{294}}  & \textit{\textcolor{red}{130}} \\
		\textbf{Problematic} & \textit{\textcolor{red}{8230}} & 1128 & 812 \\
		\bottomrule
	\end{tabular}
\end{table*} %

\subsection{Improving Targeted Classification with MTL}
\label{subsec:improvement_of_target_task}
Here, we examine MTL's ability to improve the classification performance for a known target dataset (in a train-test split), by transferring knowledge from the other datasets. 

\begin{figure}[th]
	\centering	
	\includegraphics[scale=0.825]{"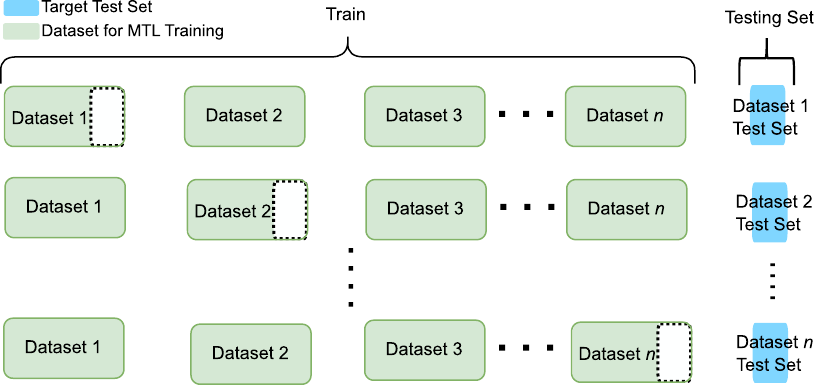"}	
	\caption{
		The problem setup for improving targeted classification. We train on all datasets while designating a target dataset, using its test set to evaluate the pipeline's ability to improve classification performance on a known dataset through transfer learning.}
	\label{fig:targeted_experiment}
\end{figure} \textbf{Experiment Setup and Evaluation.}
\cref{fig:targeted_experiment} shows the setup schema of the task.
We leverage all $n$ datasets in training; one dataset is designated as the target $d_t$ on which we aim to improve classification and the other $n-1$ are jointly leveraged by MTL.
We optimize the MTL model for each of the $n$ tasks simultaneously: on the training set of $d_t$ and the $n-1$ additional datasets.
Finally, we evaluate using $d_t$'s testing set (10\% of the dataset).
This differs from the new unseen setup used in \cref{subsec:unseen_data_classification} as the training part of the target dataset $d_t$ is not omitted.
We test against a baseline model in which we use the same model architecture as our MTL model but do not leverage any additional datasets, instead training only on $d_t$.

\begin{figure}[t]
	\centering
	\subfloat[]{
		\centering
		\begin{tabular}{llll}
			\toprule
			Target Dataset       & Baseline & MTL & p-value \\
			\midrule
			\textbf{\davidson}   & 0.6951 & 0.7450 & 0.0138 \\
			\textbf{\waseem}     & 0.7216 & 0.7823 & $<$0.0001 \\
			\textbf{\reddit}     & 0.7699 & 0.8963 & $<$0.0001 \\
			\textbf{\gab}        & 0.8621 & 0.9244 & 0.0022 \\
			\textbf{\fox}        & 0.5252 & 0.6765 & 0.0007 \\
			\textbf{\mandl}      & 0.3273 & 0.4404 & $<$0.0001 \\
			\textbf{\stormfront} & 0.6063 & 0.7400 & 0.0046 \\
			\textbf{\hateval}    & 0.6340 & 0.7526 & $<$0.0001 \\
			\textbf{\pubfig}     & 0.4534 & 0.5231 & 0.0113 \\
			\bottomrule
		\end{tabular}
		\vphantom{\includegraphics[width=0.49\columnwidth,valign=c]{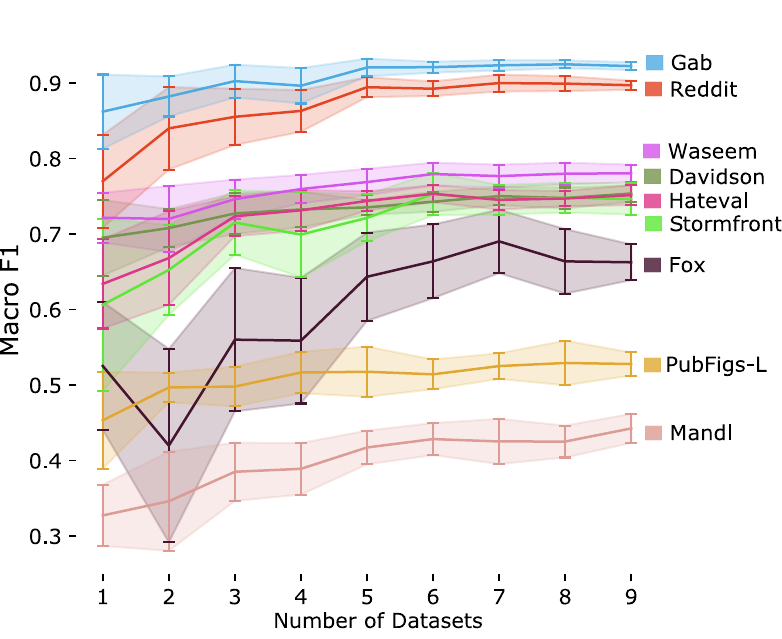}}%
		\label{tab:targeted_statistical_significance_test}
	}%
	\subfloat[]{
		\includegraphics[width=0.49\columnwidth,valign=c]{sample_n_datasets.pdf}%
		\label{fig:sample_n_datasets}%
	}%
	\caption{
		\textbf{Results of MTL targeted training.}
		(a) Mean macro F1 over 10 runs for the targeted dataset improvement task. MTL is trained with all 9 datasets while the baseline classifier is trained trained with only the target dataset. MTL's improvement over the baseline is statistically significant (p $<0.05$) for all datasets.		
		(b) Diminishing returns for increasing the datasets used in the MTL targeted training. 
		Each line represents classification on a different target dataset with the leftmost and rightmost points being equivalent to the baseline and MTL results shown in (a).
		The x-axis shows the number of datasets used in training MTL while the y-axis shows the macro F1 performance on the testing set.
		Shaded areas are the $95\%$ confidence intervals over 10 runs.
	}
	\label{fig:incresing_dataset_size_sampled}
\end{figure} 
\textbf{Outperforming the Baselines.}
We compare to the baseline model trained on only the target dataset. 
\cref{tab:targeted_statistical_significance_test} shows a comparison of the mean Macro F1 score of 10 experiment runs of the single dataset baseline compared to MTL.
To test for statistical significance, we perform Welch’s t-test \citep{welch_ttest} using the experiment runs as our samples.
We find that in all 9 datasets, our experiments show a statistically significant ($p<0.05$) improvement for MTL compared to training the model only on the target dataset.
This shows the effect of the transferring knowledge from the other datasets through the MTL framework.

We conducted an extensive literature review to identify prior studies that reported predictions on the same datasets as ours. 
\cref{tab:targeted_vs_sota} compares the performances of MTL against these works using the same performance metrics they report.
\cref{tab:targeted_vs_sota} compares the performance of MTL with the reported results of the seven identified studies, using the same performance metrics as reported by them.
The column labeled as \textit{Reported} lists the results reported in the respective studies. 
Our analysis shows that MTL outperforms the baselines consistently, with 12 wins, 1 draw, and 7 losses. 
Even in cases where MTL underperforms, it still achieves comparable performance in most cases.
As an aside (and intuitively), the targeted classification obtains significantly better results than the unseen.
For example, the unseen macro-F1 for \hateval is $0.6449$ (see \cref{tab:unseen_vs_baseline}) whereas the targeted macro-F1 is $0.7526$ (see \cref{tab:targeted_vs_sota}).

\begin{table*}
	\centering
	\caption{Confusion matrix of the \pubfig testing set for the classification head attached to a MTL unit trained on the \pubfig training set and 8 other datasets. Red italics are errors.}
	\label{tab:targeted_error_analysis}
	\begin{tabular}{llll}
		\toprule
		\multirowcell{1}{Predicted\\ Label} & \multicolumn{3}{c}{True Label} \\
		& \textbf{Neutral} & \textbf{Abuse} & \textbf{Hate} \\
		\midrule
		\textbf{Neutral}     & 1698 & \textit{\textcolor{red}{96}}  & \textit{\textcolor{red}{65}} \\
		\textbf{Abuse} & \textit{\textcolor{red}{78}} & 42 & \textit{\textcolor{red}{11}} \\
		\textbf{Hate} & \textit{\textcolor{red}{21}} & \textit{\textcolor{red}{4}} & 18 \\
		\bottomrule
	\end{tabular}
\end{table*}
 
\textbf{Diminishing Prediction Returns.}
The above results show that MTL successfully improves prediction performance for single datasets by transferring knowledge from the other datasets.
In \cref{fig:incresing_dataset_size_sampled}, we explore how the prediction performance on a target dataset $d_t$ varies with $i$, the number of additional datasets used in MTL.
For each increment of $i$, we iterate 10 times: sample without replacement $i$ datasets (non-identical to $d_t$), train MTL on the $i$ datasets and the training part of $d_t$, and report performance on the testing subset of $d_t$.

We find that the mean and variance of macro-F1 performance improve with $i$.
However, most datasets observe a diminishing returns effect, with less improvement when $i > 5$.
We posit this happens due to two phenomena.
First, some datasets transfer significantly better to others (see \cref{tab:unseen_vs_baseline}) and, as $i$ increases, so does the chance of having them in the MTL training.
Second, there is likely significant overlap in the transferable knowledge between the datasets; 
this makes newly added datasets increasingly redundant compared to datasets already used.

\textbf{Error Analysis on \pubfig.}
We perform error analysis by focusing on the performance of \pubfig classification head to gain insight into the types of errors that the classifier makes.
\cref{tab:targeted_error_analysis} shows the confusion matrix of the \pubfig classification head attached to the MTL model.
We find that the MTL model made very few mistakes when classifying neutral tweets. 
However, predictive performance on both the hate and abuse classes were poor with the majority of both classes being misclassified as neutral. 
The abuse class appears to be especially difficult to differentiate for the classifier with the most common two errors being mistaking abuse for neutral and neutral for abuse.
Performance on the hate class was poor with only $18/94$ hateful tweets correctly classified.
We attribute these results to the low number of abusive and hate speech instances, as well as differences in the definition abusive and hate speech captured in the datasets.
\pubfig exclusively features public figures who, by virtue of their prominent public profiles, are markedly less inclined to engage in overt instances of abuse or hate.
This contrasts with the other datasets which tend to feature more overt and explicit instances of hateful and abusive posts.
As such, the classifier overlooks this subtly in abusive and hateful speech in \pubfig as there is less overlap between the abusive and hate speech definitions captured by the other datasets which makes knowledge transfer through MTL less effective.

\subsection{Hate and Abusive Speech in the \pubfigfull Dataset}
\label{subsec:real_world_classification}

Here, we analyze the tweets of the 15 American public figures in \pubfigfull to gain insight into the distribution of hateful and abusive tweets and the language, topics, and targets of hate speech and abuse in their postings.

\begin{figure}[t]
	\centering
	\newcommand\myheight{0.35}
	\subfloat[]{
		\includegraphics[height=\myheight\textheight]{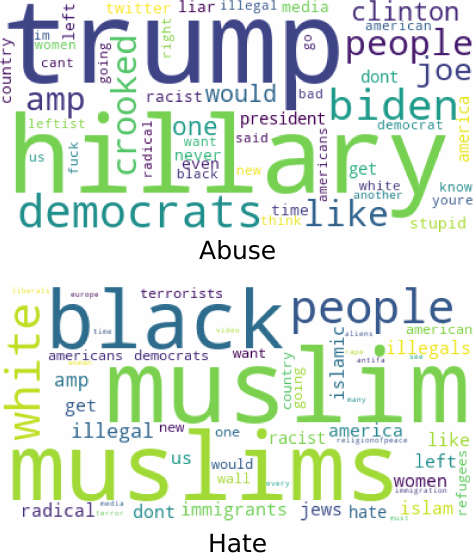}%
		\label{fig:wordmaps}%
	}%
	\subfloat[]{
		\includegraphics[height=\myheight\textheight]{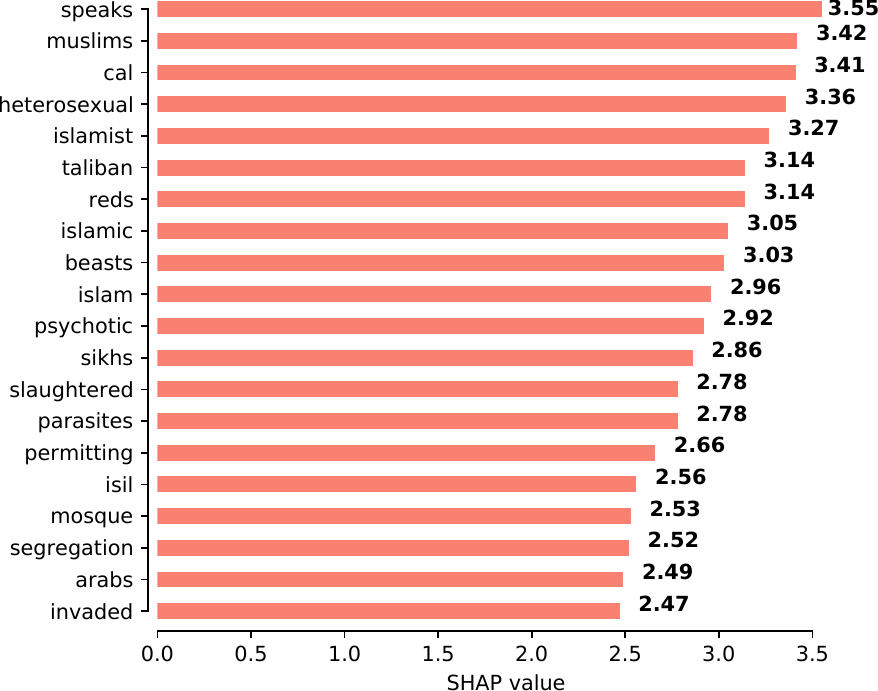}%
		\label{fig:shap_hate}%
	}%
	\caption{
		\textbf{Understanding what makes texts abusive and hateful.}
		\textbf{(a)} Most common terms in instances labeled Abuse (top) and Hate (bottom) by the MTL classifier in the \pubfigfull dataset. Larger words indicate more appearances.
		\textbf{(b)} Top 20 terms with highest SHAP values from all instances labeled ``Hate" in the \pubfigfull dataset.
	}
\end{figure} %

\begin{figure}[t]
	\centering
	\includegraphics[width=0.75\columnwidth]{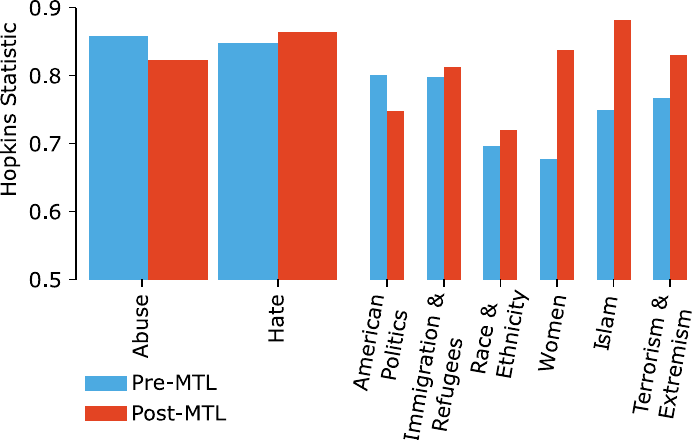}%
	\caption{
		\textbf{MTL training increases the distinctiveness of the constructed embedding space.}
		Clustering tendency measured using Hopkins Statistic value of all instances in \pubfigfull grouped by assigned label (left) and hateful instances grouped by topic keywords (right). 
	}
	\label{fig:hopkinsstatistic}
\end{figure}

\begin{figure}[t]
	\centering
	\includegraphics[width=0.75\columnwidth]{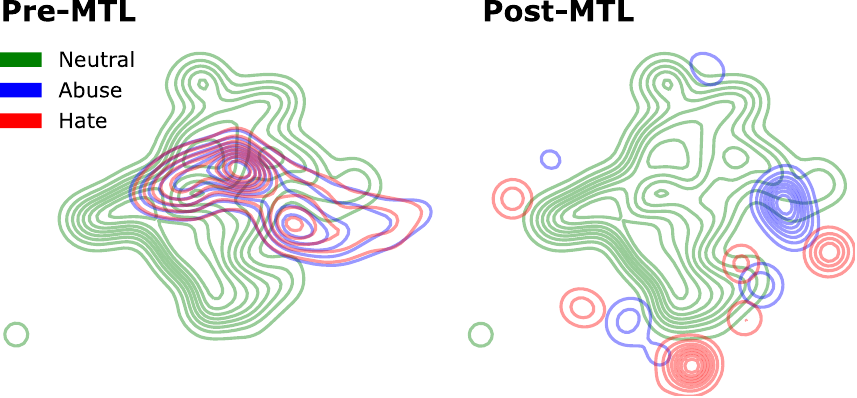}%
	\caption{
		\textbf{Abusive and hateful content becomes more localised in the post-MTL embeddings, making it more distinct and identifiable.}
		Overlayed density comparison of the embedding space constructed by pre-MTL (the off-the-shelf BERT) \textbf{(left)} and post-MTL (the MTL-tuned BERT) \textbf{(right)}.
		The embedding is projected to two dimensions using UMAP for visualization.		
		A darker color indicates a higher probability for instances of a particular label to be present at that point in the space.
	}
	\label{fig:umap_mtl_vs_unmod}
\end{figure} %

\textbf{More Hateful and Abusive Speech by Right-Leaning Figures.}
We find that right-leaning figures on Twitter produce significantly more hateful posts than left-leaning figures. 
Specifically, out of the $5299$ abusive posts, $5093$ were generated by right-leaning figures, while out of the $1133$ hateful posts, $1083$ were generated by right-leaning figures. 
These findings reinforce our earlier conclusions from profiling the \pubfig dataset, as described in \cref{subsec:pubfigs_profiling}. 
Notably, all left-leaning figures -- except the left-leaning Democrat politician IO -- had fewer than ten hateful posts. 
Upon manual inspection of IO's tweets classified as Abuse and Hate, we observed that many addressed topics related to Muslims and Islam, which are common themes in Islamophobic content. 
IO is a practicing Muslim and often tweets regarding the topic.
We posit that some of these classifications may be false positives, as the model may conflate Islam-related content with Islamophobic content due to the occurrence of similar terms in hateful posts.

\textbf{Topics and Targets of Hate Speech.}
We explore the main topics and targets of hateful and abusive speech by examining the most commonly occurring words in tweets labeled as hateful and abusive across all figures. 
\cref{fig:wordmaps} shows the word clouds of the top 100 most common words for posts classified as Abuse (top) and Hate (bottom) by the MTL classifier in the \pubfigfull dataset.
The most frequent abuse and hate words were related to six topics: Islam, immigrants and refugees, race and ethnicity, women, terrorism and extremism, and American politics. 

The top terms from instances labeled as hateful were mainly concerned with the first five topics, while those labeled as abusive mainly related to American Politics.
The latter mostly relate to individuals in American politics -- such as Donald Trump, Hillary Clinton, and Joe Biden -- whereas the hatefully labeled instances have broader targets.
The democrats Hillary Clinton and Joe Biden are the most common targets of abuse, mainly from right-leaning figures who generate most of the abusive tweets. 
As \pubfigfull covers both the 2016 and 2020 elections, it follows that the Democratic nominees are the most visible.

\textbf{What Makes A Tweet Hateful?}
We apply SHAP \citep{shap} analysis to uncover the terms the classifier deems vital for determining whether a post is hateful. 
\cref{fig:shap_hate} shows the top 20 terms with the highest absolute SHAP values. 
None of these top terms exhibit a negative SHAP value, indicating that no term strongly contributes to a post being labeled as ``not hateful''.
Furthermore, the top keywords with the highest absolute SHAP values differ from the most common terms, with indicator terms such as ``beasts", ``parasites" and ``reds" emerging as highly important in determining whether a post is hateful. 
Terms related to terrorism -- such as ``isil'' and ``taliban'' -- also had high SHAP values;
a qualitative inspection reveals that they relate mostly to immigrants and refugees, Islam, race and ethnicity.
Our findings suggest that the most vulnerable groups to hate speech are Muslims, immigrants and refugees, people of color, and women. 
At the same time, American politicians are the primary targets of abusive speech. 
Such findings highlight the need to better distinguish between Islam-related and Islamophobic content.

\textbf{MTL Training Increases Distinctness of Hate Spech Embedding.}
We investigate the impact of the MTL training on the generated embedding space.
The embeddings are large vectorial representations generated by the BERT language model for each tweet, on which downstream tasks (such as hate speech detection) can be performed.
We compare the embedding representations constructed by BERT before and after MTL tuning -- dubbed \emph{pre-MTL} and \emph{post-MTL}, respectively.
Post-MTL refers to the embeddings generated by the BERT model after fine-tuning in the targeted training setup described in the \textbf{Setup} of \mbox{\cref{subsec:improvement_of_target_task}}.
We aim to determine whether MTL improves embedding utility by examining how distinct hateful and abusive posts are in each embedding representation. 
In post-MTL, the distinction between hateful, abusive and harmless posts should increase, improving hate speech detection performance.

We evaluate distinctness using the Hopkins Statistic -- a measure for clustering tendency, i.e., how likely are instances with similar properties to be located close together in the embedding space, forming clusters.
The Hopkins Statistic compares, for each point, the nearest neighbors between the original dataset and a random dataset generated with the same size and dimension as the original. 
A value closer to $1$ indicates a strong clustering tendency, while a value closer to $0.5$ indicates a random distribution with no clustering tendency.
\cref{fig:hopkinsstatistic} plots the Hopkins Statistic for abusive and hateful tweets (left) and for each of the identified six topics of problematic speech (right) (see \textbf{Topics and Targets of Hate Speech}) using the pre- and post-MTL embeddings.
Post-MTL increases the clustering tendency for hateful tweets and slightly decreases it for abusive tweets. 
We posit this to be because all datasets used in training relate to hate speech, while only three (out of nine) datasets explicitly differentiate abusive (but not hateful) speech.
This causes abusive instances to become less distinct.
Zooming in on the topics of hate speech, we find that five (of the six) topics display an increased clustering tendency.
The clustering tendency increases substantially for two topics: \textit{Islam} and \textit{Women}.
Islamophobia and misogyny are prominent facets of hate speech, especially among the far-right~\citep{Sian2018}.
The increased clustering tendency shows that the improvement in hate speech detection may be due to BERT better distinguishing Islamophobic and misogynistic posts in its post-MTL embedding space.

We also visually inspect the distinctness of hateful and abusive posts.
We use UMAP \citep{umap} -- a widely used dimensionality reduction technique -- to project the embedding space into two dimensions.
\cref{fig:umap_mtl_vs_unmod} shows the two-dimensional density plots of the three populations of posts in \pubfigfull: neutral (green), abusive (blue) and hateful (red).
The left panel shows the pre-MTL embedding, and the right panel the post-MTL.
Pre-MTL achieves minimal separation between the two problematic classes, with the two distributions overlapping each other and the neutral class.
These are signs of low distinctiveness.
Post-MTL significantly increases the separation between hateful and abusive instances, now clustered into several dense areas.
This may be due to specific patterns within the textual content, such as common terms or similar topics.
Interestingly, the neutral distribution remains unchanged and maintains the same shape as pre-MTL.

\section{Conclusion and Future Work}
\label{sec:conclusion_and_future_works}
This work addresses the problem of poor generalizability in hate speech detection models when evaluated on new datasets. 
We propose a Multi-task Learning (MTL) pipeline that leverages multiple datasets to construct a more encompassing representation of hate to improve generalization.
Our results show that our method outperforms existing works when evaluating new unseen datasets and is comparable with state-of-the-art ones for improving performance on a known dataset.

Furthermore, we contribute a machine-labeled dataset of the online Twitter postings of American Public Political figures and a human-labeled subset. 
We apply the MTL classifier to machine-label the more than 300 thousand tweets in our dataset.
We investigate the patterns of usage of abusive and hateful speech by public political figures and the effects of MTL training on the shared embedding space representation.
We find that right-leaning figures produce significantly more abusive and hateful content than left-leaning figures, with the majority of problematic content centered around six topics. 
We also find that MTL training increases the distinctiveness of hateful and abusive speech from neutral speech.

\textbf{Limitations and Future work}
Several directions for future work can be explored based on the findings of this study. 
Firstly, current classification on new domains involves binarizing dataset class labels for unseen classification using label mapping, which can limit the specificity of results. 
A direction for further extension is increasing the specificity beyond the ``problematic vs. harmless'' binary problem. 
As our results show that MTL training increases the distinctiveness of hate in the shared embedding space, improving the specificity regarding the targets of hate speech could be a possible avenue for future work.

Secondly, future work can focus on incorporating additional context into the MTL framework. 
In this work, we process only textual data from a single posting and discard the accompanying context, such as the conversation chain and attached media (images or videos). 
Incorporating this additional context into the learning framework could improve the model's predictive power.

Thirdly, pre-trained transformer models have seen significant advancement in recent years.
This work uses a shared BERT model as its pooling point. 
Future work can explore using different and more recent models such as DeBERTa, XLM-RoBERTa, and GPT-based models over BERT.

Fourthly, this work identified several topics and targets of hate in the online discourse of public figures. 
In particular, we identified Muslims, women, and immigrants and refugees as targets of hate. 
Future research aims to analyze the impact of the detected hateful tweets on public discourse and how figures interact with such hateful speech, whether by following existing discourse or starting new discussions.

Finally, while the unseen classification results outperform the baselines and existing works, the raw F1 value remains relatively low. 
The MTL pipeline trains on several datasets simultaneously, forcing the shared model to learn representations useful to classify all datasets used, reducing the bias introduced from each dataset. 
This operates under the assumption that there is no significant overlap in the biases of each dataset, which may or may not be true in practice. 
Multiple datasets with shared biases may, in fact, further exacerbate the problem.
The exact definition of hate learned by the MTL and single dataset baselines is unknown due to the black-box nature of the model. 
Therefore, examining dataset characteristics and exploring explainability methods could provide insight into the learned definitions and the characteristics that make text hateful. 
Explicitly studying the types of biases and their overlap between datasets is a direction for future work.

\subsection*{Research Ethics and IRB approval}
Before performing any human labeling, we underwent an ethics review process by our university's Institutional Review Board (IRB) Human Ethics Committee.
The committee assessed ethical standards, guidelines, and relevant experts regarding potential ethical concerns and risks.
We subsequently conducted a comprehensive review of our project and its procedures, considering worker consent, privacy, and data security issues. 
Based on the IRB review, our project complies with ethical research standards (approval number: ETH22-7031).
\section*{Acknowledgements}
This research was supported by an Australian Government Research Training Program (RTP) Scholarship, by the Commonwealth of Australia (represented by the Defence Science and Technology Group) through a Defence Science Partnerships Agreement, and by the Australian Department of Home Affairs.
This work has been partially supported by the National Science Centre, Poland (Project No. 2021/41/B/HS6/02798).
This research was undertaken with the assistance of resources and services from the National Computational Infrastructure (NCI), which is supported by the Australian Government.
\bibliographystyle{ACM-Reference-Format}

\clearpage
\appendix

\noindent This document accompanies the submission \textit{\titlename}.
The information in this document complements the submission and is presented here for completeness reasons.
It is not required to understand the main paper or reproduce the results.
\nocite{Zhang2020,Wu2019,Rizoiu2011}

\section{Mechanical Turk Annotations}
Amazon Mechanical Turk operates in HITs (Human Intelligence Tasks). A HIT represents a single self-contained task that can be completed by workers to obtain a reward. In the context of our annotation task, each hit consisted of assigning 10 labels to 10 independent tweets from public figures. Each HIT was annotated by 8 unique workers and workers were paid \$0.1 USD per HIT completed. Workers in Amazon Mechanical Turk are allowed to do as many and as few HITs as they so choose. 
Workers were also given warnings that the content of the tweets they encounter may contain hateful and offensive speech, and were advised to cease work if they feel distressed by the content.

\begin{figure}[th]
	\centering
	\includegraphics[width=1\textwidth]{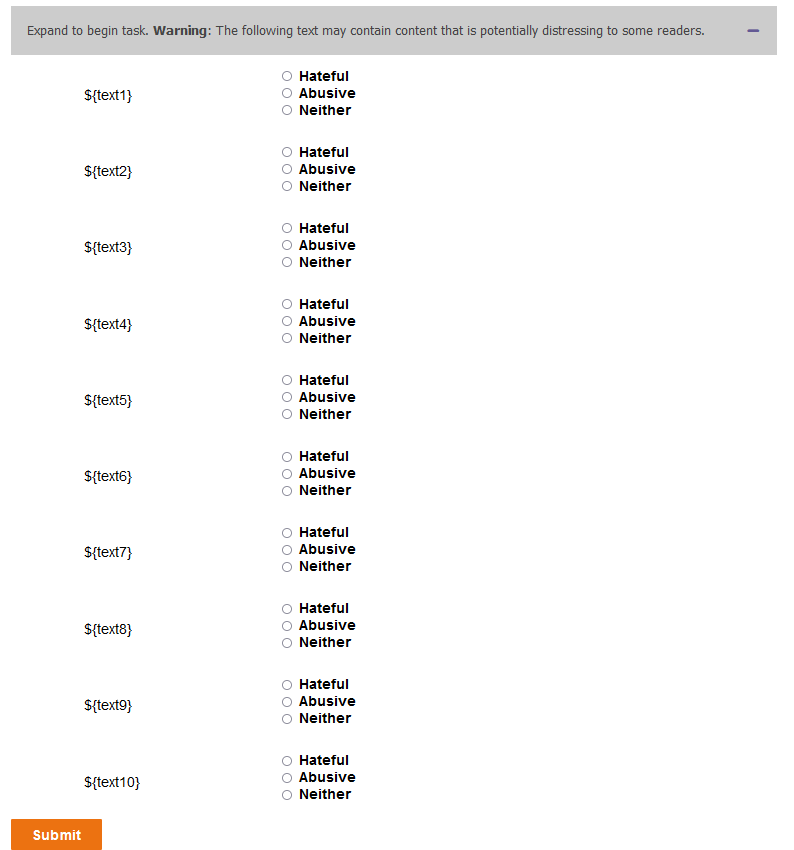}%
	\caption{Screenshot of web interface for labeling of the Tweets provided to Amazon Mechanical Turk workers.}
	\label{fig:mturk_interface}
\end{figure}

The annotation process is as follows: In each HIT, workers are asked to view the tweet text and assign one of either ``neutral", ``abusive", or ``hate'' as the label for each of the 10 tweets. The definitions for each of the labels are given at the start of the page with examples of correct annotations for each label. Screenshots of the web interface are shown in \cref{fig:mturk_interface}.

\begin{figure}[th]
	\centering
	\includegraphics[width=1\textwidth]{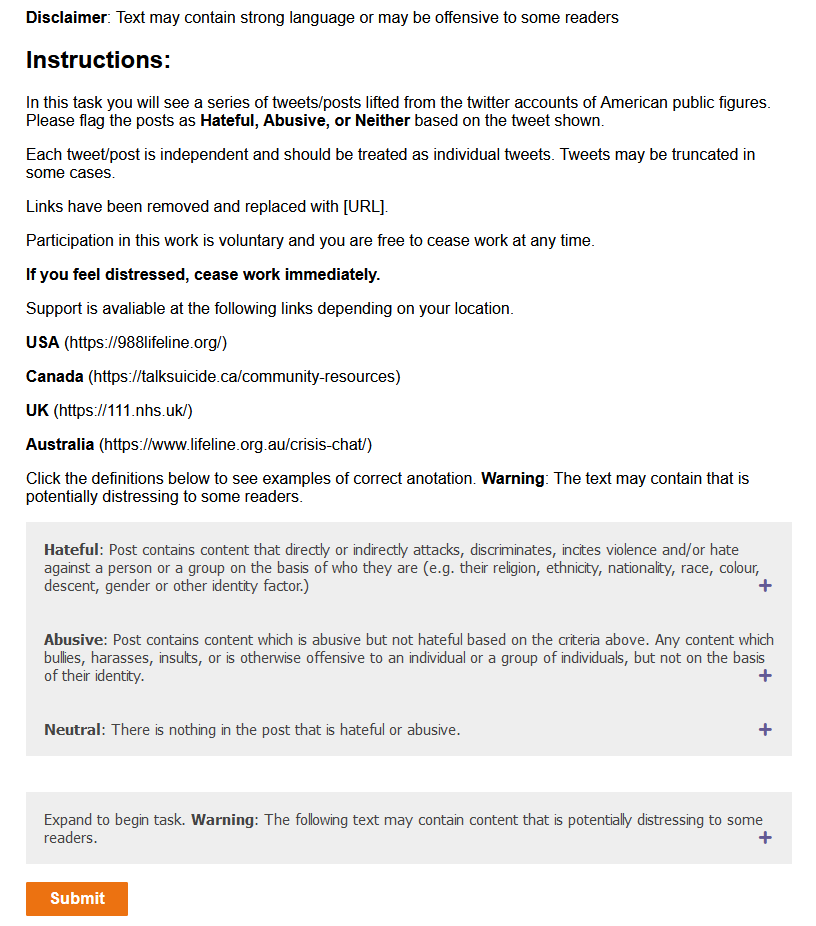}%
	\caption{Screenshot of the task instructions provided to Amazon Mechanical Turk workers.}
	\label{fig:mturk_instructions}
\end{figure}

Workers could only submit the HIT if they have completed all 10 of the tweets in the HIT. Several conditions were set on the workers in order to improve the quality of the completed work. We wanted workers whom were competent in the English language and were knowledgeable of social and political contexts in the tweets. As such, only workers with more than 5000 completed and accepted HITs, an overall approval rating of over 98\%, and residing in either the UK, the United States, Canada, or Australia were allowed to work on our task. HITs were run in batches each consisting of 50 to 100 HITs at a time over the period from April to May 2022.

\section{When MTL generalization hurts performances}
\cref{tab:unseen_vs_baseline} shows that MTL-MV and MTL-NCH underperform on the \mandl and \waseem datasets, compared to the single dataset baselines. 
The poor performance is especially significant on the \waseem dataset, with macro-F1 below all of the single dataset baselines. 
We posit that this underperformance is due to differences in the labeling criteria of the \waseem and \mandl datasets. 
The \waseem data set differs from the other datasets used in this work as it was not constructed with the goal of hate speech classification in mind, but rather to quantify the agreement between amateur and expert annotators. 
The problematic classes in the \waseem dataset also differ from the other datasets. 
The other datasets focus more broadly on hateful and offensive language while \waseem's ``Racism'' and ``Sexism'' instead focus on very specific facets of problematic speech.
Due to this, we posit that the poor performance can be attributed to the representation constructed by the MTL models being too generalized for the specific labeling criteria used in the \waseem dataset. 
Another potential factor is the high number of false positives, which \citet{Waseem-2016-racist} acknowledges. 
\citet{Yuan2023} also uncovers biases in the data set, such as a cluster of tweets that start with ``I'm not sexist, but...'' being labeled as sexist regardless of the continuation, creating many false positives. 
Underperformance on the \mandl dataset is not as significant, with only the \davidson, \pubfig, and \hateval baselines outperforming MTL-NCH. 
The performance difference between the \hateval baseline and MTL-NCH is relatively insignificant (less than 0.01) while a larger difference exists between MTL-NCH to the \davidson and \pubfig baselines.
These two datasets have similar labeling criteria to \mandl in that they explicitly differentiate between hateful speech and offensive/abusive speech. 
As the majority of datasets do not explicitly make this differentiation between the facets, the definition learnt by MTL may therefore be less vivid about the separation between hateful speech and offensive/abusive speech, causing MTL-NCH to underperform compared to the \davidson and \pubfig baselines.

\begin{table}[]
	\label{tab:topic_keywords}
	\caption{Keywords for each of the topics identified by examining the most commonly occurring words in posts flagged as hate or abuse in \pubfigfull}
	\begin{tabular}{lp{12.5cm}}
		\toprule
		Topic   & Keywords\\
		\midrule
		Islam & muslim, muslims, islamic, islam, religionofpeace \\
		Immigration \& Refugees & migrants, migrant, aliens, alien, illegal, illegals, refugees, refugee, immigrants, immigrant, immigration\\
		Race \& Ethnicity & racist, black, blacks, asian, asians, jew, jews\\
		Women & woman, women, girl, girls\\
		Terrorism \& Extremism & hamas, terror, terrorism, terrorist, terrorists, radical\\
		US Politics & hillary, trump, antifa, liberals, democrats, democrat, biden, obama, republican, republicans, president, tcot, tlot, gop, leftist\\
		\bottomrule
	\end{tabular}
\end{table} %

\begin{table*}[t]
    \caption{Ruzicka~\citep{ruzicka1958anwendung} similarity between unigrams of problematic (Red) and non-problematic (green) instances in each dataset. Results rounded to 4 d.p.}
    \label{tab:ruzika_unigrams}
    \begin{tabular}{lllllllp{1cm}ll}
    	\toprule
    	\textbf{}               & \textbf{\davidson}        & \textbf{\waseem}          & \textbf{\reddit}          & \textbf{\gab}             & \textbf{\fox}             & \textbf{\mandl}           & \textbf{\stormfront}      & \textbf{\hateval}       & \textbf{\pubfig}   \\ 
    	\midrule
    	\textbf{\davidson}       &                                & \cellcolor[HTML]{BCE3C8}0.2078 & \cellcolor[HTML]{F4F9F8}0.0531 & \cellcolor[HTML]{EAF5EF}0.0817 & \cellcolor[HTML]{C1E5CC}0.1941 & \cellcolor[HTML]{B7E1C4}0.2214 & \cellcolor[HTML]{BCE2C8}0.2082 & \cellcolor[HTML]{C0E4CB}0.1972 & \cellcolor[HTML]{E0F1E7}0.1083 \\
    	\textbf{\waseem}         & \cellcolor[HTML]{FBD4D7}0.1472 &                                & \cellcolor[HTML]{CFEAD8}0.1557 & \cellcolor[HTML]{B6E0C2}0.2265 & \cellcolor[HTML]{E4F3EA}0.0982 & \cellcolor[HTML]{C5E6D0}0.1837 & \cellcolor[HTML]{8ACE9D}0.3464 & \cellcolor[HTML]{95D3A6}0.3169 & \cellcolor[HTML]{ABDBB9}0.2564 \\
    	\textbf{\reddit}         & \cellcolor[HTML]{FAA0A3}0.3052 & \cellcolor[HTML]{FBBEC0}0.2152 &                                & \cellcolor[HTML]{63BE7B}0.4546 & \cellcolor[HTML]{FCFCFF}0.0297 & \cellcolor[HTML]{F0F7F5}0.0647 & \cellcolor[HTML]{CEEAD7}0.1593 & \cellcolor[HTML]{D0EAD9}0.1540 & \cellcolor[HTML]{B5E0C2}0.2273 \\
    	\textbf{\gab}            & \cellcolor[HTML]{FA9EA0}0.3123 & \cellcolor[HTML]{FBD4D7}0.1456 & \cellcolor[HTML]{F8696B}0.4731 &                                & \cellcolor[HTML]{F6FAFA}0.0474 & \cellcolor[HTML]{E2F2E9}0.1023 & \cellcolor[HTML]{AEDDBC}0.2469 & \cellcolor[HTML]{B0DEBE}0.2415 & \cellcolor[HTML]{81CA95}0.3729 \\
    	\textbf{\fox}            & \cellcolor[HTML]{FCFCFF}0.0224 & \cellcolor[HTML]{FCE7EA}0.0879 & \cellcolor[HTML]{FCF7F9}0.0407 & \cellcolor[HTML]{FCFBFE}0.0258 &                                & \cellcolor[HTML]{C8E7D2}0.1766 & \cellcolor[HTML]{DBEFE2}0.1233 & \cellcolor[HTML]{DEF0E5}0.1153 & \cellcolor[HTML]{F2F8F6}0.0589 \\
    	\textbf{\mandl}          & \cellcolor[HTML]{FCECEF}0.0729 & \cellcolor[HTML]{FBBDC0}0.2165 & \cellcolor[HTML]{FCDCDF}0.1205 & \cellcolor[HTML]{FCE8EB}0.0840 & \cellcolor[HTML]{FBD7DA}0.1379 &                                & \cellcolor[HTML]{BDE3C9}0.2050 & \cellcolor[HTML]{B3DFC0}0.2343 & \cellcolor[HTML]{D7EDE0}0.1327 \\
    	\textbf{\stormfront}     & \cellcolor[HTML]{FCF4F7}0.0481 & \cellcolor[HTML]{FBD0D3}0.1590 & \cellcolor[HTML]{FCEBEE}0.0764 & \cellcolor[HTML]{FCF3F6}0.0528 & \cellcolor[HTML]{FAAEB0}0.2642 & \cellcolor[HTML]{FBBBBD}0.2247 &                                & \cellcolor[HTML]{85CC99}0.3605 & \cellcolor[HTML]{AADBB9}0.2576 \\
    	\textbf{\hateval}        & \cellcolor[HTML]{FBC1C3}0.2052 & \cellcolor[HTML]{FAA9AB}0.2784 & \cellcolor[HTML]{FAACAE}0.2694 & \cellcolor[HTML]{FBBFC2}0.2096 & \cellcolor[HTML]{FCECEF}0.0723 & \cellcolor[HTML]{FBC0C2}0.2087 & \cellcolor[HTML]{FCDADC}0.1296 &                                & \cellcolor[HTML]{A1D8B1}0.2828 \\
    	\textbf{\pubfig} & \cellcolor[HTML]{FCEEF1}0.0662 & \cellcolor[HTML]{FBB8BB}0.2322 & \cellcolor[HTML]{FCE1E4}0.1059 & \cellcolor[HTML]{FCEBED}0.0774 & \cellcolor[HTML]{FBC8CB}0.1833 & \cellcolor[HTML]{FAA5A8}0.2900 & \cellcolor[HTML]{FAADB0}0.2647 & \cellcolor[HTML]{FBC2C4}0.2030 &                                \\
    	\bottomrule
    \end{tabular}
\end{table*}

\end{document}